\documentclass[runningheads]{llncs}

\usepackage{graphicx}
\usepackage{color}
\usepackage{subcaption}
\usepackage[symbol]{footmisc}
\usepackage[noend,linesnumbered]{algorithm2e} 
\usepackage{tabularx}

\usepackage{xcolor}

\newcommand{\tablfeaturea}{0.25}
\newcommand{\tablfeatureb}{0.75}

\usepackage[width=122mm,left=12mm,paperwidth=146mm,height=193mm,top=12mm,paperheight=217mm]{geometry}
\begin{document}
% \linenumbers
\pagestyle{headings}
\mainmatter

\title{Learning 3D Shapes as Multi-Layered Height-maps using 2D Convolutional Networks} 
% Replace with your title

\titlerunning{Multi-Layered Height-maps for 3D shape processing}
% Replace with a meaningful short version of your title

\authorrunning{Sarkar et al.}
% Replace with shorter version of the author list. If there are more authors than fits a line, please use A. Author et al.

%\author{Authors bla bla }

\author{
Kripasindhu Sarkar$^{1,2}$\and 
Basavaraj Hampiholi$^{2}$\and \\
Kiran Varanasi$^{1}$\and 
Didier Stricker$^{1,2}$
}

%Please write out author names in full in the paper, i.e. full given and family names. 
%If any authors have names that can be parsed into FirstName LastName in multiple ways, please include the correct parsing, in a comment to the volume editors:
%\index{Lastnames, Firstnames}
%(Do not uncomment it, because you may introduce extra index items if you do that, we will use scripts for introducing index entries...)

\institute{$^1$DFKI Kaiserslautern \quad
	$^2$Technische Universit{\"a}t Kaiserslautern\\
	\email{ \{kripasindhu.sarkar,basavaraj.hampiholi,\\
	kiran.varanasi,didier.stricker\}@dfki.de}
}

\maketitle

\begin{abstract}
We present a novel global representation of 3D shapes, suitable for the application of 2D CNNs. We represent 3D shapes as multi-layered height-maps (MLH) where at each grid location, we store multiple instances of height maps, thereby representing 3D shape detail that is hidden behind several layers of occlusion. We provide a novel view merging method for combining view dependent information (Eg. MLH descriptors) from multiple views. Because of the ability of using 2D CNNs, our method is highly memory efficient in terms of input resolution compared to the voxel based input. Together with MLH descriptors and our multi view merging, we achieve the state-of-the-art result in classification on ModelNet dataset.
\end{abstract}

\section{Introduction}
Over the last few years, Convolutional Neural Networks (CNNs) have completely dominated in solving vision based problems in 2D images achieving the state-of the-art-results in various domains \cite{Krizhevsky2012,Simonyan2014,He2015a,Girshick2014,Girshick2015,Ren2015,Long2015a,He2017,Radford2015,Ledig2016,Pathak2016}. 
%Over the last few years, Convolutional Neural Networks (CNNs) have completely dominated in solving vision based problems in 2D images, achieving the state of the art results in various domains, such as classification \cite{Krizhevsky2012,Simonyan2014,He2015a}, detection \cite{Girshick2014,Girshick2015,Ren2015}, segmentation \cite{Long2015a,He2017} and generation \cite{Radford2015,Ledig2016,Pathak2016}. 
These methods are motivated through the large amount of work in designing core network architectures, such as GoogLeNet \cite{Szegedy2015a}, ResNet \cite{He2015a}, InceptionV3/V4 \cite{Szegedy2015} etc. because of a) the ease of performing convolution operation on the 2D image grids and b) the availability of large scale labelled image databases such as ImageNet \cite{Russakovsky2015}.

However, applying the ideas from these powerful CNNs on 3D shapes is not straightforward, as transcribing the shapes into a common parameterization/description is a necessary first step for the application of CNN. The simplest descriptor - the voxel occupancy grid - makes it possible in theory to apply analogous 3D networks of 2D images (VGG, ResNet, Inception, etc.) to the 3D voxel representation. In practice, it is not feasible as the memory and computation grows `cubically' with resolution of the voxel representation, making it difficult to perform research in designing core network for 3D shapes. Thus, the existing voxel based 3D networks are limited to low input resolutions (on the order of $32^3$) \cite{Maturana2015,Wu2015,Su2015,Brock2016}. Other methods for 3D specific tasks are the invention of new geometric representation or network architecture targated for 3D shapes \cite{Riegler2017a,Qi2016,Klokov2017}, and the use of appearance based approaches of using rendered images \cite{Sfikas2017,Johns2016,Su2015}.
Appearance based methods are by design not appropriate for geometry based tasks such as shape generation, reconstruction, etc., though they are excellent choices for appearance based tasks such as classification and retrieval. 

In this paper, we present a novel geometry centric descriptor for 3D shapes, suitable for the application of 2D CNNs. We represent 3D shapes as multi-layered height-maps. At each grid location, we store multiple instances of height-maps, thereby representing 3D shape detail that is hidden behind several layers of occlusion. Using this parameterization, that is intuitive and simple to construct, we learn 3D shapes using 2D convolutional neural network models and show state-of-the-art classification result on the ModelNet dataset \cite{Wu2015}.
Our descriptor provides the following advantages: 1) It is geometry centric, making  it appropriate for solving both appearance and geometry based tasks 2) It enables the use of well-investigated 2D CNNs in the context of 3D shapes, which is not possible in voxel based representation and other new 3D architectures; and the ability of taking advantage of pretrained 2D CNNs trained using large scale image data. 3) As a consequence, it provides a highly memory efficient CNN architecture for 3D shapes which is comparable to that of OctNet \cite{Riegler2017a}, while being similar in performance.

The multi-layered height-map (MLH) representation is generic and suitable to any 3D shape, irrespective of topology and volumetric density in shape representation. It does not need a pre-estimation of 3D mesh structure, and can be computed directly on point clouds. 
%Using the Modelnet dataset, we show that a small number of layers is sufficient to represent 3D shapes with many parts. 
Our work is in contrast to more shape-aware parameterizations which require the knowledge of the 3D mesh topology of the shapes, which can then be used to create a mesh quadrangulation, or learning an intrinsic shape description in the Eigenspace of the mesh Laplacian \cite{Masci2015,Sarkar2018}. 
%Although these shape-aware parameterizations are powerful and represent the geometry more efficiently, they cannot be quickly adapted into a common parameterization that is valid across multiple shapes. 
Our MLH parameterization is suited for learning the shape features in a large dataset of diverse 3D shapes. In this sense, it is comparable to 3D voxel grids, but without the associated memory overhead. 
%With the publication, we will make MLH descriptors of ModelNet40 as well as our trained network available to public. 
Our contributions are the following:
\vspace{-0.2cm} 
\begin{itemize}
  \item We propose a novel multi-layered height-map (MLH) based global representation for 3D shapes suitable for 2D CNNs for various tasks.
  \item We propose a novel multi-view merging technique for CNNs involving different input branches to combine information from multiple sources of an instance into a single compact descriptor.
  \item We present state-of-the-art result on ModelNet benchmark \cite{Wu2015} for classification using our multi-view CNN and MLH descriptors.
\end{itemize}

The following section describes the related work. Section \ref{sec:mlhdescriptors} explains in detail the multilayered height-map based features for 3D shapes and simple 2D CNNs for the problem of classification. We present in Section \ref{sec:viewmerging} our Multi-view CNN architecture for combining the global features along different views. 
We follow that with the experiments section evaluating different design choices.

\section{Related Work}
\noindent
\textbf{Core 2D convolution networks} 
%2D convolutional networks have been applied accross several 2D domanins such as classification\cite{Krizhevsky2012,Simonyan2014,He2015a}, detection \cite{Girshick2014,Girshick2015,Ren2015}, segmentation \cite{Long2015a,He2017} and generation \cite{Radford2015,Ledig2016,Pathak2016}. 
AlexNet\cite{Krizhevsky2012} was the first deep CNN model trained in GPU for the task of classification, and is still often used as the base model or feature extractors for performing other tasks. Other famous models which are often used as base CNN are VGG \cite{Simonyan2014}, GoogLeNet \cite{Szegedy2015a}, ResNet \cite{He2015a}, InceptionV3/V4 \cite{Szegedy2015}. VGG is a simple network which uses a series of small convolution filters of size $3\times3$ followed by fully connected layers. GoogLeNet and InceptionV3/V4 models provide deeper networks with computational efficiency containing efficient `inception' modules. ResNet on the other hand uses only $3\times3$ convolution with residual connections. We use the 16 layered VGG \cite{Simonyan2014} as our base CNN model because of its simplicity.
%\vspace{-0.3cm}

\noindent
\textbf{3D convolution networks on voxel grid}
Voxel sampling is the method where a 3D shape is represented as a binary occupancy grid in a 3D voxel grid. Wu et al.\cite{Wu2015} uses deep 3D CNN for voxelized shapes of resolution $30^3$ and provides the classification benchmark dataset of ModelNet40 and ModelNet10.
This work is followed by VoxNet which uses voxels of resolution $32^3$ \cite{Maturana2015}. Recently, network elements from 2D CNNs such as inception modules and residual connections have been integrated in 3D CNNs which gave a huge performance gain over the traditional 3D CNNs \cite{Brock2016}. Because of the fundamental problem of memory overhead associated with 3D networks, the input size was restricted to $32^3$. Fine-grained analysis specific to shape classification and 3D CNN have been performed in \cite{Qi2016a,Sedaghat2016} making them the top performers in shape classification. In contrast to voxel gird, we use our multi-layer descriptors and use 2D CNN and perform better both in terms of accuracy and computation overhead in the task of shape classification in ModelNet benchmark.
%\vspace{-0.3cm}

\noindent
\textbf{View-dependent rendering methods}
Image-view based methods take some sort of virtual snapshots (rendering or depth image) of the shape and then design a 2D CNN architecture to solve the task of classification. Their contributions are combination of a novel feature descriptors based on rendering \cite{Sfikas2017,Johns2016,Su2015}, and novel changes in the network architecture for the purpose of classification based on appearance  \cite{Wang2017a}. As explained in Sections \ref{sec:otherrepresentation} and \ref{sec:otherresults}, our representation with 1 layer performs similar for the task of classification in comparison to the image-view based methods.
%\vspace{-0.3cm}

\noindent
\textbf{2D slices} Gomez-Donoso et al. \cite{Gomez-Donoso2017} represents shape by `2D slices' - the binary occupancy information along the cross section of the shape at a fixed height. A multi-view CNN architecture is then developed to feed 3 such slices (across the 3 canonical axes) for classification. In contrast to this work, (1) our MLH descriptor have \textit{k} height values (k $\approx 5$) from the reference grid, and therefore informative enough to be used as a descriptor even for single view CNN, (2) our descriptor is generative (full shape outline can be generated - Section \ref{sec:resGAN}) and holds promise towards solving other geometry centric tasks.

\noindent
\textbf{Specialized networks for 3D}
More recently, there has been a serious effort to have alternative ways of applying CNNs in 3D data such as OctNet \cite{Riegler2017a} and Kd-tree network \cite{Klokov2017}. 
%PointNet takes unstructured 3D points as input and gets a global feature by using max pool as a symmetrical function on the output of MLP (multi-layer perceptron) on individual points. 
Kd-tree network uses Kd tree as the underlying data structure and learns a representation of the input for solving various tasks, providing an excellent alternative of CNN on 3D data. OctNet on the other hand, uses a compact version of voxel based
representation where only the occupied grids are stored in an
octree instead of the entire voxel grid. It has similar computational
power as the voxel based CNNs while being extremely memory efficient enabling 3D CNNs with $256^3$ input. We show that our one view descriptor of resolution 256 and a simple 2D CNN performs similar to OctNet in terms of classification accuracy and memory requirements. 
%\vspace{-0.3cm}

\noindent
\textbf{Unordered point clouds and patches} 
It is possible to sample the 3D shape to a finite number of 3D points and collect their XYZ coordinates into a 1D vector. This representation is compact, but it has no implicit spatial ordering that aligns with the real world. Achlioptas et al. \cite{Achlioptas2017} in a recent submission uses such represenation to generate 3D shapes and also achieve good accuracy in ModelNet10. PointNet \cite{Qi2016} is another such network that takes unstructured 3D points as input and gets a global feature by using max pool as a symmetrical function on the output of multi-layer perceptron on individual points. Our method is conceptually different as it respects the actual spatial ordering of points in the 3D space. 
Sarkar et al.~\cite{Sarkar2017b,Sarkar2018} learn from a dataset of unordered 3D patches, which are detected and oriented using a quadrangulation approach. They represent the spatial ordering at the patch level, but not at the global context of the 3D shape like our method. Further, our method does not require such a prior quadrangulation step.

%\section{Approach}

\begin{figure*}[t!]
\begin{subfigure}{0.5\linewidth}
  \centering
  \includegraphics[width=\textwidth]{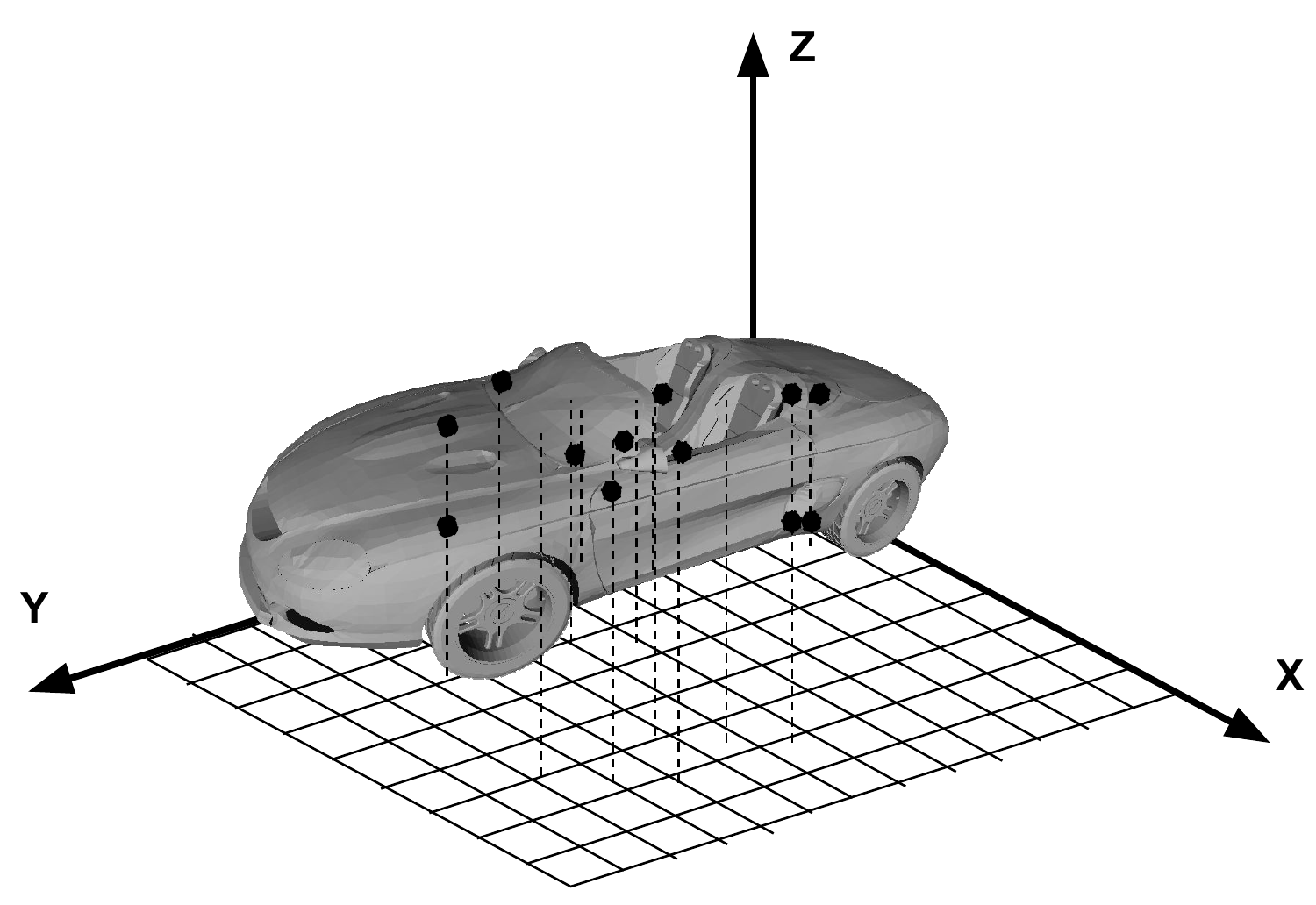}
\end{subfigure}%
\begin{subfigure}{0.5\linewidth}
  \centering
  \includegraphics[width=\textwidth]{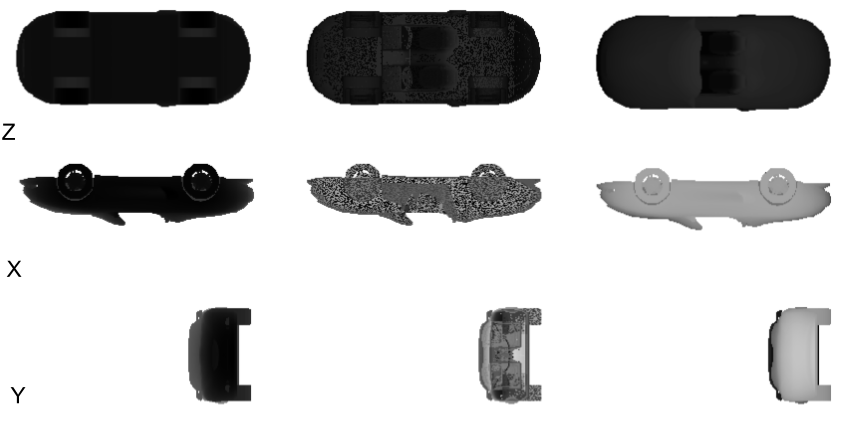}
\end{subfigure}% 
  \caption{(Left) Multi-layered height-map descriptors for a shape with the view along Z. (Right) Visualization of the corresponding descriptor with $k = 3$ from 3 different views of X, Y and Z.}
  \label{fig:mlhdescriptors} 
  %\vspace{-0.5cm}
\end{figure*}

\section{Multi-Layered Height-map descriptors}
\label{sec:mlhdescriptors}
%Multi-layered representation is a geometry centric global description of 3D models which can be used for various tasks such as shape classification, retrieval, generation etc. Its key feature is the ease of the use of 2D CNNs for the aforementioned tasks. 
MLH descriptor represents a 3D shape with multiple instance of `height-map' from a discrete reference grid depicting multiple surface layers. In comparison to voxel occupancy grid structure, where each voxel bin stores the model occupancy information, our representation stores a list of $k$ `heights' or displacement maps in each bin of the 2D reference grid. The idea is to consider $k$ sample height values of the entire cross-section of the shape intersecting or falling along the bins of the 2D grid. For implementation of this idea, we first convert the shape into a point cloud and process on them as explained in Algorithm \ref{alg:mlhalg}. 

\RestyleAlgo{boxruled}
\begin{algorithm}[t!]
%\scriptsize
%\tiny
%\small
\SetKwInput{Notation}{Notation}
\SetKwInput{Input}{Input}
\SetKwInput{Output}{Output}
\SetKwInput{InputRansac}{Params plane fitting}
\SetKwInput{Initialise}{Initialise}
\DontPrintSemicolon
 \Notation{$A_{[i,j,k, \ldots]}$ denotes the element of $A$ at index $i,j,k, \ldots$}
 \Input{shape - $S$, resolution - $N$, number of layers - $k$, direction $\hat{n}$}
 \Output{MLH descriptor M of dimension $(N \times N \times k)$}
 \Initialise{$M \gets full($Inf$)$}
	Orient $S$ using $\hat{n}$ and scale it to contain in unit bounding box. \\
	Densely sample points in $S$ to get the point-cloud $C$.\\
	Place an $N \times N$ square grid of unit length on the X-Y plane of the boundig box. \\
	Project $C$ in the grid and collect the z-coordinates (height values) in the bins.\\
  \ForEach{bin $(p,q) \in \{1, \ldots, N\}$}{
  	\tcp{let $P_{pq}$ be the set of height values of the points falling in the bin $(p,q)$}
  	\If{ $P_{pq}$ is not empty}{
	when $k>1$, $M_{[p,q, i]} \gets ((i-1)/(k-1) * 100)^{th}$ percentile of $P_{pq}$ for each $i \in \{1, \ldots, k\}$\\
	otherwise, $M_{[p,q, i]} \gets 0^{th}$ percentile $P_{pq}$ (or minimum of $P_{pq}$).  	
  	}
 } 
 \caption{Computation of MLH descriptors}
 \label{alg:mlhalg}

\end{algorithm} 

%Formally, given a shape S in a unit bounding box, a discritized plane (2D grid) with normal $\hat{n}$, resolution $(N \times N)$, size (1 x 1) and center aligned to the center of the bounding box, we consider all the rays from the $N^2$ bins of the grid perpendicular to itself. For each such ray we find the point of intersections $P_b$ to the shape. We then choose $k$ points in $P_b$ based on some strategy and represent them by its distance to the plane for all the bins to get a $(N \times N \times k)$ descriptor for the entire shape representing our multi-layered height-map (MLH) descriptor. Point cloud of the corresponding shape can be generated given its descriptors $D$ by $\cup_{x, y, i} (x/N, y/N, D(x, y, i))$. We call the $(N \times N)$ 2D grid  at $i^{th}$ index/height as layer $i$. Therefore our representation is composed of $k$ 2D arrays of dimension $(N \times N)$. The method is explained in Figure \ref{fig:mlhdescriptors}.

%\paragraph{\textbf{Discussion}} 

The empty bins with no surface intersection are represented by a value slightly higher than the maximum possible height (Inf = 1.2, or surface with infinite height), for differentiating them from the valid height values. The points are uniformly and densely sampled from the shape so that we get atleast $k$ points in a bin when it is occupied (Step 2). We take percentile values for different layers (in comparison to other sampling strategy - Eg. `slices' at equidistant locations, minimum $k$ values etc.). This preserves the semantics for the 1$^{st}$ and k$^{th}$ layers as the bottom and top surfaces respectively. The layers between them represent the intermediate shape information hidden from outside.
%\paragraph{\textbf{Layer Strategies}} There could be several ways of choosing $k$ points out of all the surface points $P_b$ along the bin $b$ of the discrete grid. The strategy of choosing the points for a layer is important based on the application. One strategy is to sort them based on the height value, and take the $k$ minimum points. This representation produces $k$ different sheet like layers, with the $i^{th}$ layer denoting the $i^{th}$ distant surface from the reference plane. The empty bins with no surface intersection are represented by a value slightly higher than the maximum possible height, Inf (= 1.2) for all layers, for differentiating them from the normal heights (or surface with infinite height). Layers with bins having less than $k$ surface points are filled with Inf.  While this approach may be good for generating shapes, semantics about the layers (except the 1st layer) are lost. 

%Other strategy is to consider points at k intermittent layers of the shape with the $1^{st}$ and $k^{th}$ layers representing the bottom and top surfaces. This strategy preserves the semantics for the 1st and kth layers being the bottom and top surfaces respectively, and the layers in between denoting the intermittent shape information. We find this representation to be the best for the task of classification and all the experiments except otherwise mentioned uses this representation.

\textbf{View direction} The MLH representation is dependent on the plane normal $\hat{n}$ - the direction along which the height-map is computed - making it a view based descriptor. We call this \textit{view direction} in the subsequent sections. More on the choice of the view directions are covered in the subsequent sections.

%When the shapes are oriented along consistent axes, we can choose one of their canonical axes as the view directions for features computation. 
%More on the view directions and its combination with its results are explained in details in the subsequent sections.

\subsection{Comparison to other shape representations}
\label{sec:otherrepresentation}
\subsubsection{Voxel sampling} In contrast to 3D voxel occupancy grid, our representation stores distance from the reference plane in the view direction instead of the binary occupancy. Since continuous distance is more precise than the discretized occupancy bins, our representation provides more information along the view direction, provided the number of surfaces falling on a bin is less than $k$. This case is already satisfied for most of the cases with $k$ = 5, except for the surfaces parallel (or near parallel) to the view directions. 
%Having MLH computed from the three canonical directions X, Y and Z of the shape enables all the surfaces to have atleast one direction not perpendicular to them. 
Therefore MLH in general, is more expressive than voxel occupancy grid with a good chosen direction while being less in memory ($N^3$ vs $kN^2$). 

%\vspace{-0.3cm}
\subsubsection{Depth images} With $k = 1$, our feature descriptor reduces to depth image representation of a shape with orthographic projection (instead of perspective projection of the depth camera).
%\vspace{-0.3cm}
\subsubsection{Rendered images} 
%Rendered images of shapes are also used heavily in the vision community for the task of classification and retrieval of shapes \cite{Su2015}. The idea is to render the shape from different views, and use the rendered images for solving the problem classification and retrieval. 
Even though a rendered image is dependent on shading model, geometric 3D features (corners and edges) appear as a result of perspective projection of the 3D model to the 2D plane of the image. Therefore our representation with $k = 1$ is similar in nature to the rendered images with orthogonal projection. This premise is supported by the similarity of our result of classification accuracy in ModelNet40 with $k = 1$, to the popular technique of MVCNN \cite{Su2015} which uses rendered images.

\subsection{Classification networks}
\label{sec:classificationnetworks}
Due to the fact that MLH descriptors are multi channel 2D grids, we can directly apply any feed forward 2D CNN with a classification loss (eg. cross entropy loss or SVM loss) for classification. In the simplest form, we use the view of one consistent direction in our MLH features. For incorporating different views, we can treat each view as different training instance and take the sum of all the views at the testing time. We can also treat the views separately and have a merging technique coming from different views. We discuss this in detail in the next section.

This also enables us to use popular 2D CNNs such as AlexNet, VGG, ResNet etc (with $k$ input channels instead of 3 for images) for 3D shape related tasks. These popular networks have been trained on ImageNet database consisting of millions of images. Since the number of instances in 3D databases are much less than that of 2D databases (12K in ModelNet40 compared to 1.2M in ImageNet), we proceed with a fine-tuning strategy for our 3D shape classification. This is meaningful as our feature is analogous to $k$ channel images (instead of 3 for real images). Since the networks trained on images meaningfully capture the image details at various levels, we expect them to provide good results when fine-tuned on `image like' inputs. 
%This premise is reinforced by the result.

%Except for classifier of DCGAN for shape generation, we used VGG16 as our base model with $k$ input channel in all of our experiments because of its simplicity and small size, though the results might be better by training with newer networks such as ResNet and InceptionV4 etc.
%\vspace{-0.3cm}
%\noindent
\textbf{Initialization strategy}
The weights from a network trained on 2D images can be used to initialize our corresponding network for fine-tuning, except for the first layer where the number of input channels (with k = 5), and hence the weights of the convolutional layer, do not match. In this case, we can either initialize the first layer randomly (and proceed through fine-tuning with other layers properly initialized from a pre-trained network), or copy part of the weights for the first layer. 
%In particular for the network VGG16, the dimension of the weights for the first layer for our representation is $5 \times (3 \times 3) \times 64$ denoting 64 filters of $5 \times (3 \times 3)$ dimensions, compared to $3 \times (3 \times 3) \times 64$ for images. 
We copy the weights of image-based network to the 1st, 3rd and 5th channel of the weights in the network for MLH. For the remaining 2 layers, we use the average of the weights of the 3 channels of image-based network and copy them to that of MLH network. 
%This has been done with the premise that 1) first convolutional layers identifies the lower level features for images - such as edges and corners, which are important in the context of MLH descriptors, and 2) different channels in images (R, G and B) are similar in nature in terms of low level features. And therefore, the weight of the convolutional kernel corresponding to each layer are similar in nature. 3) 1st, 3rd and 5th layers are the most important layers in the 5 layered MLH descriptors (due to its strong semantics). 
This initialization strategy provides us with an improvement of approximately 0.1\% in test accuracy (over randomly initialized first layer) in all the experiments with ModelNet40/10.

%\mytodo{provide visualization of kernel}.
%\mytodo{provide result improvements}.

\section{Multi-view architecture} 
\label{sec:viewmerging}
The MLH representation is a view dependent descriptor, and thus a well designed network for a problem should consider features computed from multiple views. We therefore need a good strategy for choosing views, and a technique to merge information coming from different views to have a single descriptor. This section discusses the design choices. 
%\vspace{-0.3cm}

\subsection{Choice view directions} 
%As discussed before, our representation provides more information along the view direction. This case is satisfied with a well chosen number of layers - $k$, except for the surfaces parallel (or near parallel) to the view directions. 
Having MLH computed from the three canonical directions X, Y and Z of the shape, enables all the surfaces to have atleast one direction not perpendicular to them. When the 3D data is not axes aligned, we just use the three canonical axes as view direction and proceed through MVCNN \cite{Su2015} like architecture to combine the three views. %When the data is axes aligned, we make some changes to the existing view merging methods by the following way.

%MLH features computed from three orthogonal directions enables all the surfaces to have atleast one direction perpendicular to them. Most of the online repositories and online databases for 3D shapes (ShapeNet, ModelNet etc.) provide axes aligned shapes. Eg - for a car, the gravity direction is always Z, front as Y and the side as X in ModelNet/ShapeNet dataset. In these scenario we consider 3 views along X, Y and Z for computation of such features for a more complete representation. If shape-aware axes can be selected in the context of other datasets e.g, through medial axis computation, it is possible to adapt our method accordingly, by choosing them as view directions. 

%\vspace{-0.3cm}
\subsection{View merging requirements for aligned data} 

\textbf{Aligned data} Most of the online repositories and online databases for 3D shapes (ShapeNet, ModelNet etc.) provide axes aligned shapes. Eg - for a car, the gravity direction is always Z, front as Y and the side as X in ModelNet/ShapeNet dataset. This important meta-information has been exploited successfully for various tasks implicitly or explicitly \cite{Sfikas2017,Sedaghat2016,Johns2016,Soltani2017}. In the availability of such aligned data, our MLH features from X, Y and Z directions have more specific meaning. We design a multi-view architecture specifically to exploit this information. Note that if shape-aware axes can be selected in the context of other datasets e.g, through medial axis computation, it is possible to adapt our method accordingly, by choosing them as view directions. 

%\noindent
\textbf{MVCNN} Merging information coming from different views is addressed in MVCNN \cite{Su2015}, which has been the state-of-the-art method for shape classification for a long time. We first explain MVCNN, and discuss its merits and demerits before proposing our own solution for merging different views.

In MVCNN, a given shape is first rendered from several consistent directions, which forms the input to the task specific CNN. 
Each rendered image is passed through a CNN branch separately. All the branches in this part of the network share the same parameters. For merging the outputs from different branches, an element-wise maximum operation across the different activation volume is taken. This is followed by the second part of the network which consists of Fully Connected (FC) layers followed by a loss layer for training. Key elements of this design are 1) shared weights for all the branches, and 2) element-wise max pooling for merging the output from different views. Given the nature of the problem, we identify the following disadvantages in this design. 

\begin{enumerate}
\item The element-wise max merging operation makes the network give more importance to one of the input views, and discard the information of the others. 
\item The element-wise max commutative operation makes the merged output agnostic to the order of the input views, which is not justified except for random view directions. 
\item Weight sharing in the branches makes the network give less importance to the semantics of the views. The network gets updated in the same way among all the views even with different semantics of input to them.
\end{enumerate}

%3D shapes in the public inventories (3D Warehouse, ShapeNet etc.) and datasets (ModelNet) are axes aligned, which is an important meta-information. 
Presence of aligned 3D datasets makes it important to process different views separately, and explicitly differentiate the output coming from different views. 

\begin{figure*}[t!]
  \includegraphics[width=\textwidth]{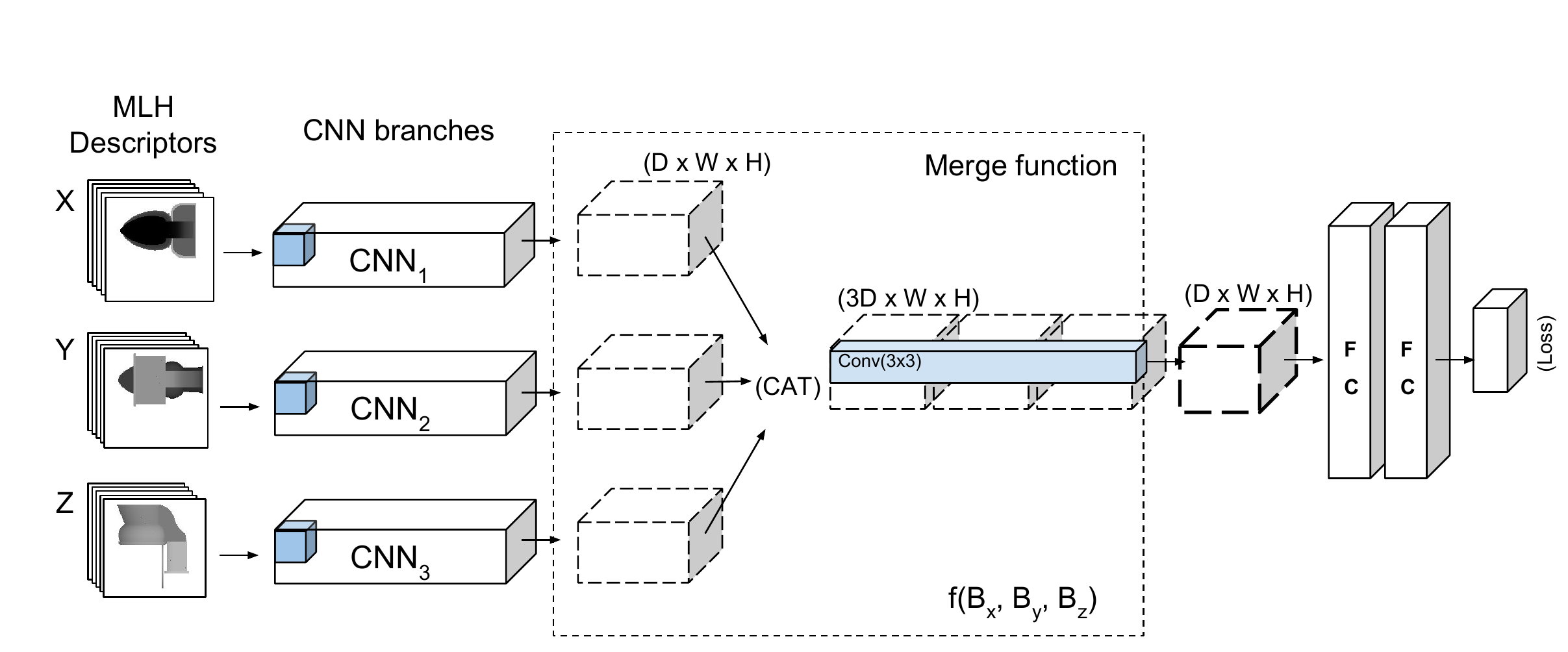}
  \caption{Overview of the view-merging operation in our Multi-view architecture.}
  \label{fig:mvcnn} 
  %\vspace{-0.5cm}
\end{figure*}

%\vspace{-0.3cm}
\subsubsection{Network design formulation}
Our multiview network takes input $(X_1,..., X_N)$ coming from N different views. Each of them gets passed through $N$ CNN branches $(c_1,..., c_N)$, giving the output $(B_1,..., B_N) = (c_1(X_1),..., c_N(X_N))$. We then perform a merging operation $f$ on the outputs of the branches to get the final output $f(B_1,..., B_N)$. This is followed by fully connected (FC) layers and eventually by a loss function during the training. Based on analysis on MVCNN, we have the following design choices in our network.

\begin{enumerate}
\item \textbf{Independent view branches} The network branch for each view should be independent of each other. i.e., the CNN branches $c_i()$ have different weights.

\item \textbf{Non-commutative merging operation} To explicitly differentiate the merged output before the application of the FC layers, we use a non-commutative merge operation. i.e., the function $f: B^n \rightarrow M$ is non-commutative.
\end{enumerate}

\subsection{Multi-View classification network for MLH descriptors}
\label{sec:mergingoperation}
Our Multi-view network takes 3 MLH feature descriptors as input from X, Y and Z directions. The CNN branches $c_i$s are  simple feed forward 2D convolutional networks. We use one of the popular 2D CNN architectures trained on ImageNet, as we can use the trained weights to initialize our model. 
%In particular, we use the pre-trained VGG16 (till its last Conv layer) network as our CNN branches in all of our experiments related to classification \mytodo{experiment with resnet50, inceptionv3?}. 
As mentioned above, the 3 branches do not share weights.

%\vspace{-0.5cm}
\subsubsection{Choice of non-commutative operation} Popular merging operations like max, mean and sum are all commutative, which makes the order of input irrelevant. The operation concatenation is non-commutative, but produces a large size of activation volume making it infeasible to add a subsequent FC layer (Eg. concatenating the activation of the last Conv/Pool layer of VGG16 from 3 branches and adding a FC layer result to 3*7*7*512*4096 $\approx$ 300M parameters). 

We chose the operation - \textit{convolution followed by concatenation} as our choice of non-commutative operation. This reduces the size of the concatenated output back to the initial amount and enables adding a subsequent FC layer to the network. Specifically, we concatenate the activation volumes of the 3 branches with dimension $D \times W \times H$ along the axis of depth, to get a concatenated volume of  $3D \times W \times H$. We then apply $3 \times 3$ convolution with $D$ filters to get the volume back to $D \times W \times H$ before applying the subsequent FC layer. These trainable weights of the convolution filters along the entire concatenated volume make the operation non-commutative. However, the network can learn to make these weights non-commutative or commutative based on the type of input during learning. The design is illustrated in Figure \ref{fig:mvcnn}.

\section{Experiments}
%We perform experiments to evaluate the overall aspects of the multi-layered height-map descriptors and the merging technique. 
%We use the task of shape classification to quantify our results for our different design choices. 

\subsection{General settings for shape classification}

\textbf{Dataset} We use the popular datasets of ModelNet40 and ModelNet10 in \cite{Wu2015} for evaluating our shape classification results. ModelNet40 contains approximately 12k shapes from 40 categories, whereas ModelNet10, a subset of ModelNet40,  contains approximately 5k different shapes. These datasets come with their training and testing splits ($\approx$ 10k and 2.5k shapes for ModelNet40; 4k and 1k for ModelNet10). %Both of the datasets have been extensively used by various work on 3D shape descriptors for evaluating their design. 
We computed MLH feature descriptors of dimension $256 \times 256 \times 5$ and performed no data augmentation.  

\noindent
\textbf{General Network settings} We use the VGG16 with batch normalization\cite{Simonyan2014} (without the FC layers) as our base model for both single view and 3 view merged network. More recent networks such as ResNet50 \cite{He2015a} and Inception based models \cite{Szegedy2015}, provided no improvement in results in terms of test classification accuracy, possibily due to the less number of training samples in these 3D shape datasets. These deeper network architectures may provide advantage as the 3D datasets become larger in size. We add a cross entropy loss at the end of the last FC layer and do an end-to-end training for classification.
%\vspace{-0.4cm}

\noindent
\textbf{General training settings}
We train ournetwork for 20 epochs with a batch size of 8 using SGD optimizer. The initial learning rate is set to 0.01 and is decreased by a factor of 10 after 10 epochs. Our 3 view network takes approximately 2 hours to converge with a GeForce GTX 1080 Ti GPU. 

\begin{table}[t]
\centering
\begin{tabular}{|c| c|} 
\hline
View axis & Accuracy  \\
\hline
 X & 86.91  \\ 
 Y & 86.71  \\ 
 Z & 86.91  \\ 
 \hline
\end{tabular}
\quad
\begin{tabular}{|l| c|} 
\hline
Merge settings & Accuracy  \\
\hline
a) Shared branches + max merge& 91.25   \\ 
b) Independent branches + max merge& 91.29 \\ 
c) Independent branches + cat merge & 93.11  \\ 
\hline
\end{tabular}
\caption{Classification accuracy on ModelNet40 with (Left) single view, and (Right) multiple views with different merge techniques. `cat merge' denotes our non commutative merge operation - concatenation followed by convolution.}
\label{table:parameval}
%\vspace{-1cm}
\end{table}

\subsection{Evaluation of design choices} 
We compute 5 layers ($k = 5$) MLH feature, and train single view one branch CNNs. We post the test classification accuracy for ModelNet40 in Table \ref{table:parameval} (Left). Except for the experiments with the number of layers in Section \ref{sec:reslayers}, all the experiments provided in this section use $k = 5$. Even with single view and a very simple network architecture, our classification accuracy is comparable to the popular voxel based methods. The next section provides a more detailed comparison with the state-of-the-art methods.
%\vspace{-0.3cm}
\subsubsection{View merging}
We consider the 3 canonical axes of X, Y and Z as the 3 orthogonal view directions, and perform experiments with the following branch merging operations a) MVCNN type \cite{Su2015,Qi2016a} network with shared CNN branch followed by elt max b) independent branch followed by elt max c) our design of independent branch with non-commutative merge - concatenation followed by convolution. In the last design we concatenated $512 \times 8 \times 8$ output volume of the last convolution layers of 3 VGG16 branches to get an output volume of $1536 \times 8 \times 8$. We follow this with $3 \times 3$ convolution with 512 filters (\#parameters = (1536*3*3)*512) to get the output volume back to the order of previous magnitude - $512 \times 7 \times 7$ (we reduce the dimension from 8 to 7 to properly initialize the pretrained FC layer). This operation is detailed in Section \ref{sec:mergingoperation}. The classification accuracy on ModelNet40 with the above design choices is reported in Table \ref{table:parameval} (Right). The best result is obtained with our proposed merging technique.
%\vspace{-0.3cm}
\subsubsection{Effect of fine-tuning}
One of the important features of our MLH descriptor is the fact that, even though it is a geometry based descriptor which captures geometric properties of a 3D shape, it can be easily used by pretrained image based 2D CNNs. All of our results, except otherwise stated are obtained by fine-tuning the weights of the VGG16 pretrained on ImageNet with an initialization strategy explained in Section \ref{sec:classificationnetworks}. Finetuning makes a big difference in the result in the test accuracy. For example, our 5-layer 3-view model achieves a test accuracy of \textbf{89.63} when initialized with random weights compared to \textbf{93.11} when initialized with pretrained weights, giving an relative improvement of 3.5\%. 

%\vspace{-0.5cm}
\subsubsection{Number of layers}
\label{sec:reslayers}
\begin{figure*}[t!]
\begin{subfigure}{0.4\linewidth}
  \centering
  \includegraphics[width=0.9\textwidth]{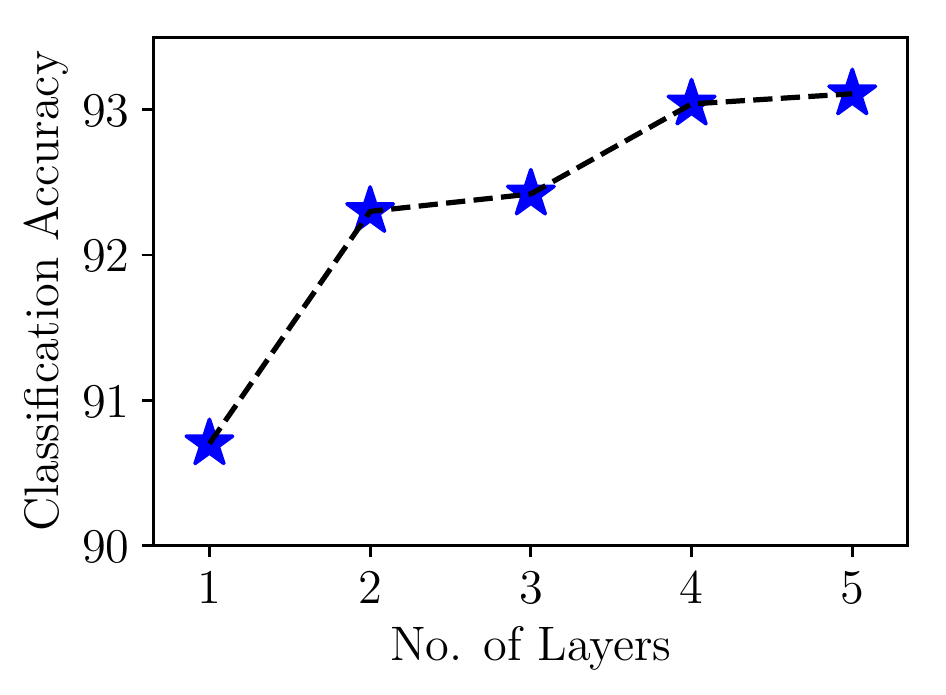}
\end{subfigure}%
\begin{subfigure}{0.6\linewidth}
  \centering
  \includegraphics[width=0.8\textwidth]{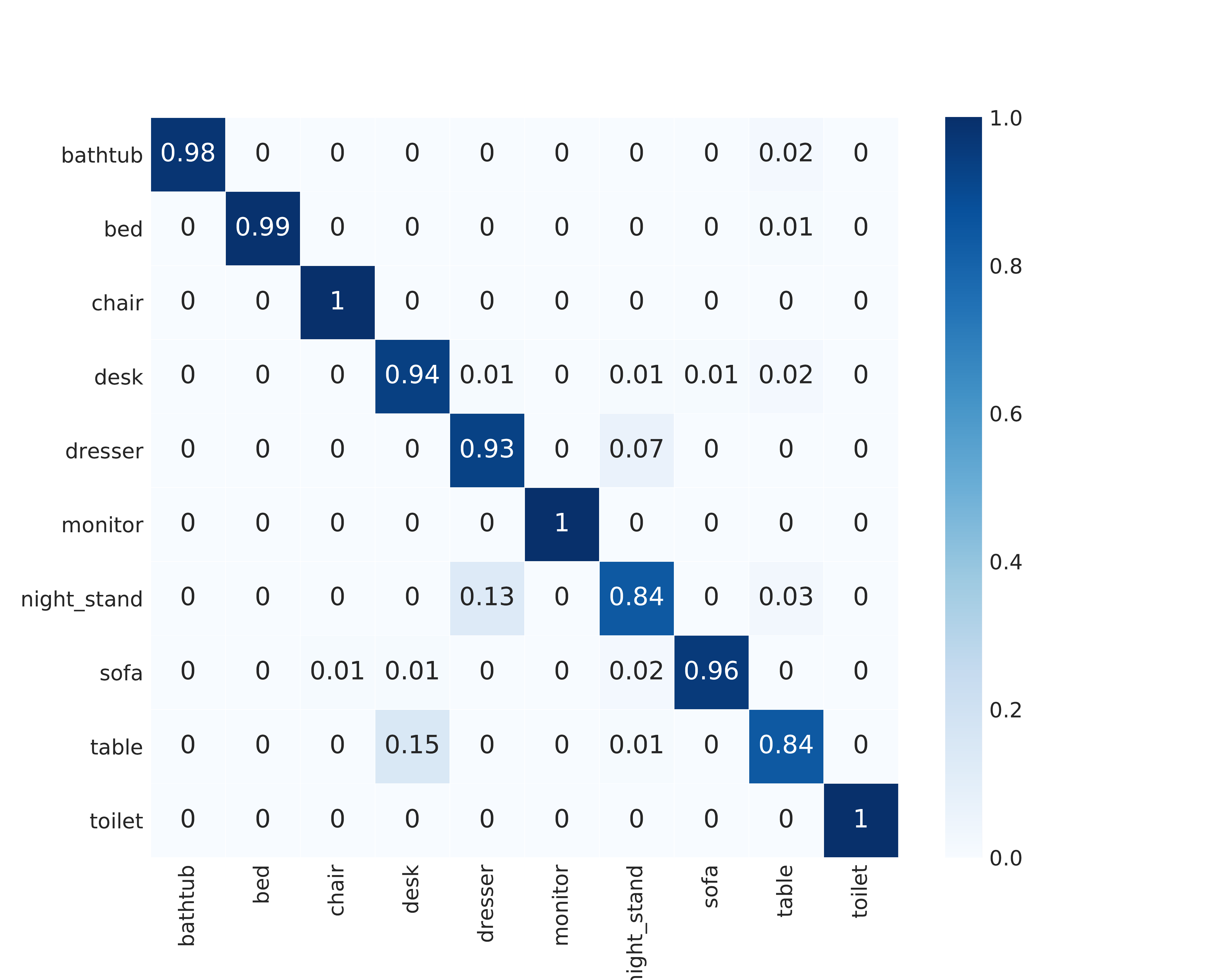}
\end{subfigure}% 
  \caption{(Left) Classification accuracy on ModelNet40 with different of number of layers. (Right) Confusion matrix of classification for ModelNet10.}
  \label{fig:nolayers} 
  %\vspace{-0.5cm}
\end{figure*}

Figure \ref{fig:nolayers} shows the effect of the number of layers in our multilayer representation. Here we use our merged 3-view models . Note that even with a single layer we achieve an accuracy higher than 90.5\% due to our new merging technique. Our 2 layer MLH descriptor already provides a good representation of the shape as it completely covers the shape outline from outside along the view direction (by taking minimum and maximum height form the grid). As expected, we see a big jump in accuracy from layer 1 to 2, followed by a slow saturation till layer 5. We therefore, use the 5 layer descriptors in all other experiments.

%\begin{figure*}[t!]
%\centering
%  \includegraphics[width=0.5\textwidth]{nolayers.pdf}
%  \caption{Classification accuracy on ModelNet40 with different of number of layers}
%  \label{fig:nolayers} 
%\end{figure*}

\begin{table}[htbp]

\begin{center}
\small
\begin{tabular}{|l|c|c|c|c|c|}

\hline
 & ModelNet40 & ModelNet10 & T converge & \# views$^a$ & Data\\ 
 & & & (ModelNet40) & \# aug inst $^b$ & aug\\\hline
 \textbf{Image based}  & \multicolumn{1}{l|}{} &  &  &  &\\ 
PANORAMA NN \cite{Sfikas2017} & 91.12 & 90.70 & 30 mins &  1$^a$& N\\ 
Wang et al \cite{Soltani2017} & \underline{\textbf{93.80}} & - & 20 hours &  12$^a$& N\\ 
Pairwise \cite{Johns2016} & 90.70 & \textbf{92.80} & - &  12$^a$& N\\ 
MVCNN \cite{Su2015} & 90.10 &  -& - &  80$^a$ & N\\ \hline
% & \multicolumn{1}{l|}{} &  &  &  \\ 
\textbf{Geometry based } & \multicolumn{1}{l|}{} &  &  &  &\\ 
Kd-Net depth 10 \cite{Klokov2017}  & 90.60 & {93.30} & 5 days &  10$^b$& Y\\ 
Kd-Net depth 15 \cite{Klokov2017}  & 91.80 & {94.00} & 16 hours &  10$^b$& Y\\ 
MVCNN- MultiRes \cite{Qi2016a} & 91.40 & - & - & 20$^a$ & Y\\ 
VRN \cite{Brock2016} & 91.33 & {93.61} & 6 days & 24$^b$ & Y\\ 
Voxception \cite{Brock2016} & 90.56 & 93.28 & 6 days & 24$^b$ & Y\\ 
ORION \cite{Sedaghat2016} & - & {93.80} & - & - & Y\\ 
Pointnet \cite{Qi2016} & 89.20 & - & - & 1$^b$ & Y\\ 
VoxNet \cite{Maturana2015} & 83.00 & - & - & 12$^b$ & Y\\ 
\underline{Ours - 3 views + cat} & \textbf{93.11} & \underline{\textbf{94.80}} & 2 hours & 3$^a$& N \\ \hline
\end{tabular}
\end{center}
\caption{Comparison of test classification accuracy on ModelNet40 and ModelNet10 dataset among all single model methods. Bold figures are the highest accuracy in the respective group. Underscore figure is the highest accuracy value among all the methods. - means information not available. \textit{T converge} denotes the Time taken for the CNN to Converge in a single GPU. Y/N means yes and no respectively.}
\label{table:soa}
\vspace{-0.8cm}
\end{table}

\subsection{Comparison with the state-of-the-art methods}
We provide a brief discussion on the different methods for shape classification before providing our comparison; to better understand the state of the art and the comparing methods.
%\vspace{-0.3cm}

%\noindent
\textbf{Image-view based methods} Image-view based methods take virtual snapshotsof the shape and then design a 2D CNN architecture to solve the task of classification. They are, in their core, appearance based methods which are excellent for solving tasks based on appearance - eg. shape classification and retrieval. They are not generative in nature (being able to generate the shape or a partial shape given the feature descriptors) and hence by design not appropriate for geometry based tasks such as shape generation, reconstruction, part segmentation etc. Because of the fact that shape classes are highly appearance centric, they contribute to the top part of the shape classification leaderboard.
%\vspace{-0.5cm}

%\noindent
\textbf{Geometry based methods} These methods use a geometry centric input, such as voxel grid, point cloud etc. and design an appropriate CNN architecture to solve the task of classification. They are geometry centric and can be used for tasks such as shape generation, part segmentation etc. 
%Our MLH representation falls in this category, though it can be reduced to appearance based descriptors as discussed in Section \ref{sec:otherrepresentation} and Section \ref{sec:otherresults}.

Table \ref{table:soa} shows the comparison of our approach to the other state of the art methods for shape classification. We perform the best compared to all the single model methods in ModelNet10, and best among all the geometry based methods in in ModelNet40. The confusion matrix for ModelNet10 is shown in Figure \ref{fig:nolayers}. Most of the missclassification comes from the similar category pairs like (\textit{table}, \textit{desk}) and (\textit{night stand}, \textit{dresser}). We also perform better than all the appearance based methods, except Wang et al \cite{Soltani2017} which performs a specialized view based clustering for the task of classification and takes 10 times longer to converge than our algorithm. It can be argued that the result of our method can be improved with more view based specialization, data augmentation and other fine-grained analysis (Eg clustering of MLH features instead of rendered images) which is not our contribution or our claim. 
%Our contribution is a novel geometric feature descriptor and a generic merging technique for combining information from different persistent sources.

\subsection{Relation with other descriptors}
\label{sec:otherresults}
%\vspace{-0.1cm}
\subsubsection{Reduction to image based models}
As described in Section \ref{sec:otherrepresentation}, our representation with 1 layer reduces to rendered image with orthogonal projection. We perform experiments with $k=1$ with similar merging setting as MVCNN and provide our results in Table \ref{table:viewreduction}. The similarity of our results verifies this hypothesis. A slightly less accuracy of our method with 1 layer and the merge settings of MVCNN is probably due to the less number of views (3 compared to 80). We see an increase in accuracy with our new merging method with 1 layer descriptors. This provides the hint that our merging method can be used by existing image based methods such as MVCNN to improve the classification accuracy.
%\vspace{-0.6cm}
\subsubsection{Comparison to Voxel based models} 
%Our representation provides advantages over voxel based representation as argued in Section \ref{sec:otherrepresentation}. 
We compare classification accuracy and memory requirements for our single view (X), 5 layer model with the results provided in OctNet paper \cite{Riegler2017a} on the ModelNet10 dataset. We chose to compare our results with this work because, a) it focuses on `pure' 3D convolutional network on voxel representation b) it isolates the effect of memory requirement on accuracy from other network design factors. On a similar note, we perform classification using a single view network using VGG16 (a simple 2D convolutional network of 16 layers - in comparison with 14 layers 3D Convnet used in OctNet256) for our MLH descriptors of dimension $256 \times 256 \times 5$. As seen in Table \ref{table:viewreduction} (Right), the result of our method using a single view is comparable to pure 3D convolution based methods on 3D voxel grid and OctNet, while being similar in the network memory requirements of OctNet.
\begin{table}[t]
\centering
\small
\begin{tabular}{|c| c|} 
\hline
Algorithms & Accuracy  \\
& ModelNet40 \\
\hline
MVCNN \cite{Su2015} (80 v) & 90.10 \\ 
{1 l + Shared + max (3 v)} & 89.78\\ 
{1 l + Ind + max (3 v) }  & 90.15  \\ 
{1 l + Ind + cat (3 v)} & 90.72  \\ 
\hline
\end{tabular}%
\quad
\begin{tabular}{|c| c| c|} 
\hline
Algorithms & Accuracy & Approx. \\
& ModelNet10 & Memory \\
\hline
Octnet 256 \cite{Riegler2017a} & 90.3 & 8 GB \protect\footnotemark \\
DenseNet 64 \cite{Riegler2017a} & 90.0 & 6 GB \protect\footnotemark\\
DenseNet 256 \cite{Riegler2017a} & - & 60 GB \\ 
\textbf{Ours 256} (1 view) & 90.97 & 8 GB \protect\footnotemark\\ 
%\textbf{Ours (single view) wo FT } & 88.99 & 110 MB \\ 
\hline
\end{tabular}
\caption{(Left) Classification accuracy on ModelNet40 with our 1 layered descriptor with 3 views and its comparison with an image based method. See Table \ref{table:parameval} for the legends. (Right) Comparison of classification accuracy and memory requirement (of batchsize = 32) of our single view model with OctNet \cite{Riegler2017a}.}
\label{table:viewreduction}
%\vspace{-0.7cm}
\end{table}

\addtocounter{footnote}{-2}
\footnotetext{Memory occupied in GeForce GTX 1080 Ti for a batchsize of 32 and approximate value inferred from the Figure 7 (a) in \cite{Riegler2017a}.}
\addtocounter{footnote}{1}
\footnotetext{Activation volume of the network in Table 5  for the batchsize of 32 and approximate value inferred from the Figure 7(a) \cite{Riegler2017a} (details in supplementary material).}
\addtocounter{footnote}{1}
\footnotetext{Activation volume of our single view network (details in supplementary material) and memory occupied in GeForce GTX 1080 Ti for a batchsize of 32.}

\subsection{Multi-view GAN on MLH descriptors}
\label{sec:resGAN}
In this set of experiments, we use the MLH features as generative feature descriptors and show that they can be applied for point cloud generation, using a DCGAN type network. We also provide a novel multi-view GAN which generates multiple views simultaneously and synchronously. We show qualitative results of the rendered point cloud of the generated shapes. 
%We clarify that we do not intend to perform an extensive evaluation on multi-view GANs which altogether merits different set of contributions with different design choices. 
Our main intention here is to verify the fact that the 2D grid of MLH descriptors (and the corresponding 2D CNNs) have sufficient geometric information and hold promise towards the application of 3D shape generation. 

We design a generative network which automatically takes care of the synchronization of multiple views by using a multiview discriminator with our merging technique. The generator takes a latent noise as input and feeds it to 3 different generating branches consisting of transposed convolutions. None of the generating branches share any weights. The design of generator not sharing any weights and a multi-view discriminator with non commutative merge operation is even more important in the context of GAN than for a simple classification network. This is because in GAN, the generator has to produce 3 different output as different views, and the discriminator has to discriminate and differentiate among the views.

\begin{figure*}[t!]
\begin{subfigure}{\linewidth}
  \centering
  \includegraphics[width=0.9\textwidth]{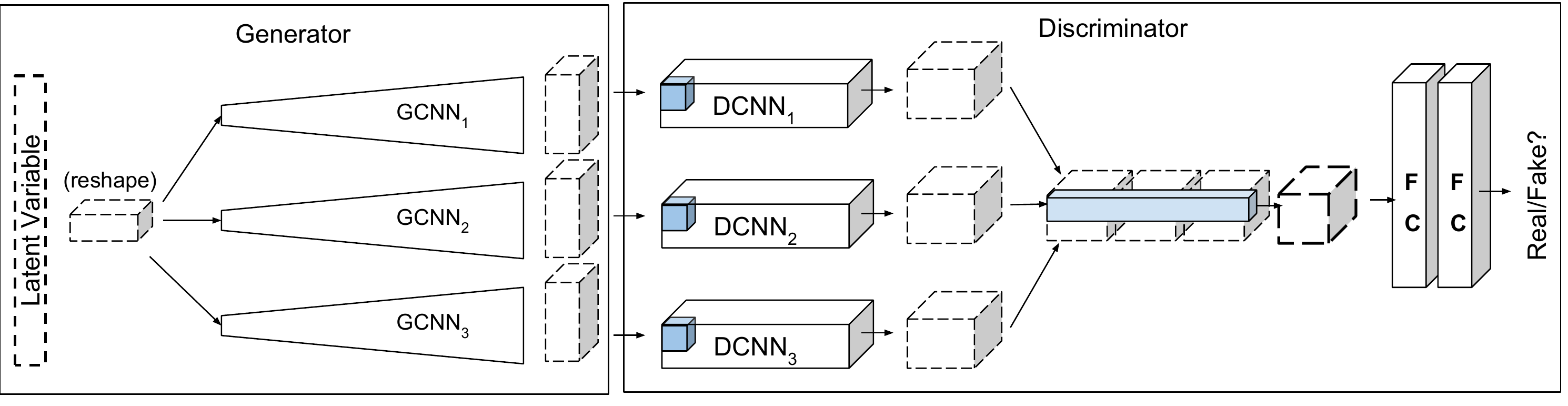}
\end{subfigure}
\begin{subfigure}{0.7\linewidth}
  \centering
  \includegraphics[width=\textwidth]{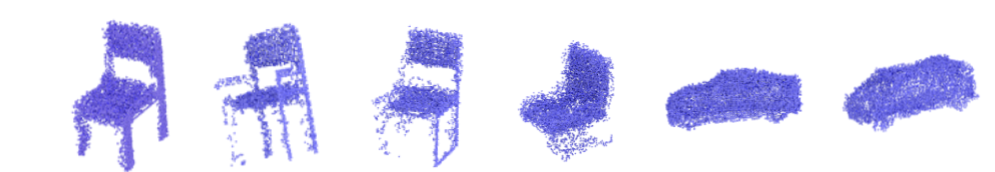}
\end{subfigure}%
\quad \quad \quad 
\begin{subfigure}{0.15\linewidth}
  \centering
  \includegraphics[width=\textwidth]{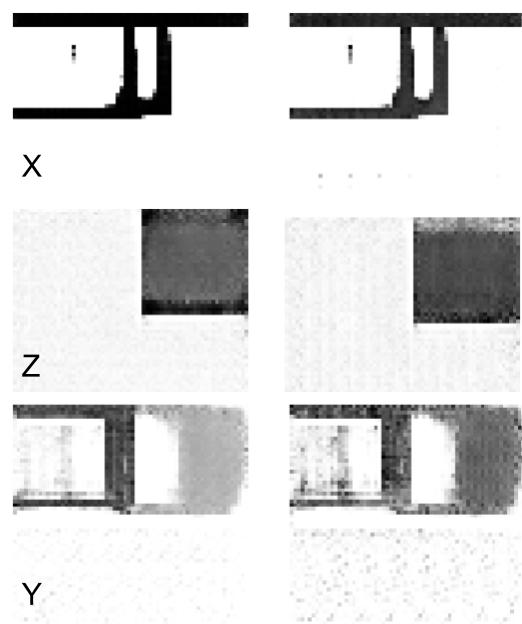}
\end{subfigure}% 
  \caption{MVDCGAN - Multiview DCGAN for MLH descriptors. (Top) The generator and discriminator architecture used in the experiments. (Bottom) Generated shapes of 3 chairs and 2 cars using the above architecture. (Bottom-Right) Visualization of the generated descriptors for one of the chair.}
  %\vspace{-0.5cm}
  \label{fig:gansamples} 
  
\end{figure*}

For generative network branches, we use 5 transposed-convolution blocks to upsample. For discriminator branches, we used 5 convolution blocks to downsample. The network details are provided in the supplementary material. We used 2 layered representation to capture the shape outline from outside. The Multiview DCGAN model together with the generated shapes are shown in Figure \ref{fig:gansamples}. We see that the generated 3D shapes show characteristic 3D features. In future work, they can be explored towards shape synthesis applications.

\section{Conclusion and future work}
In this paper, we introduced a novel 3D shape descriptor for 2D convolutional neural networks and an efficient merging technique for merging information from different views of same instance. We showed its advantages in terms of classification accuracy and memory requirements, as compared to the voxel based methods. Our method is complementary to fine-grained analysis such as view clustering (Eg. \cite{Soltani2017}) and making ensembles of classifiers, for further improving the classification accuracy. Our merging can also be used in various existing work involving merging of aligned data instances (Eg. \cite{Qi2016a,Su2015}). We also plan to perform detailed work on MV-DCGAN, both using images and MLH descriptors. We hope our MLH descriptors will provide an alternative way of 3D shape processing in the future and encourage researchers to investigate in novel 2D CNNs for solving tasks related to 3D data.

\bibliographystyle{splncs04}
\bibliography{mlmvcnn}

\appendix
\section{Introduction}
In this short document, we provide the supplementary information for the paper `Learning 3D Shapes as Multi-Layered Height-maps using 2D Convolutional Networks' - referred as \textit{Main Paper}. 

\section{Feature visualization}
Figure \ref{fig:fieature} shows the visualization of the features of a few shapes.

\section{Memory calculation}
Table \ref{table:octnet256} and \ref{table:singleview} provide the calculation of the memory in different networks, whose values are used in \textit{Table 3 (Right), Main paper}.

\section{ModelNet40 misclassified shapes}
Table \ref{table:miscl} shows some of the misclassified shapes for the ModelNet40 dataset.

\section{Network for Multi-View DCGAN}
Table \ref{table:gan} shows the detailed network architecture used in the experiments with MV-DCGAN (\textit{Section 5.5, Main paper}).

\begin{figure}[htb!]
\centering
\begin{subfigure}{\tablfeaturea\linewidth}
  \centering
  \includegraphics[width=\linewidth]{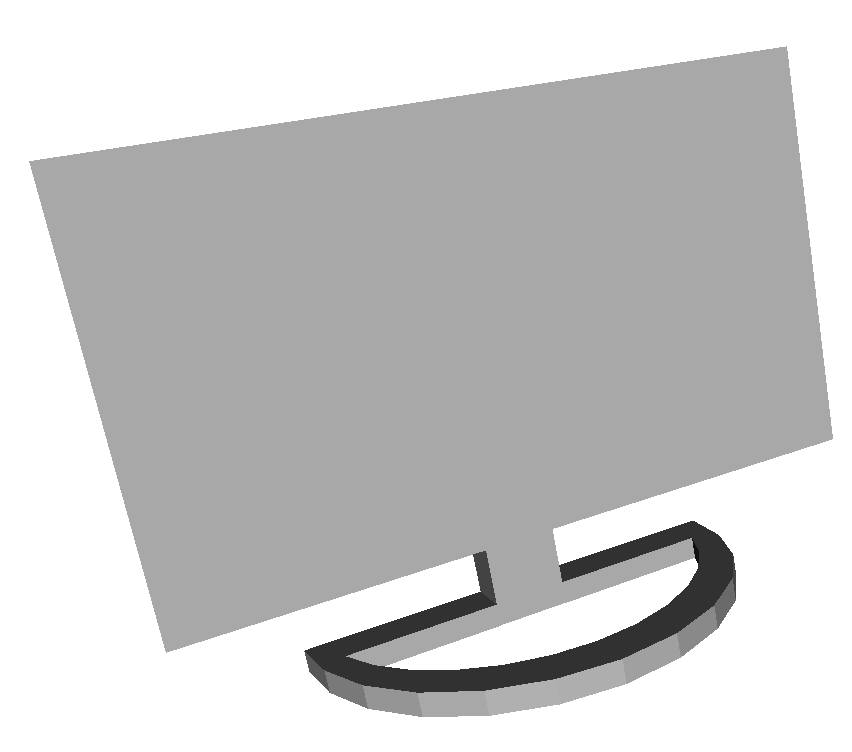}  
\end{subfigure}%
\begin{subfigure}{\tablfeatureb\linewidth}
  \centering
  \includegraphics[width=\linewidth]{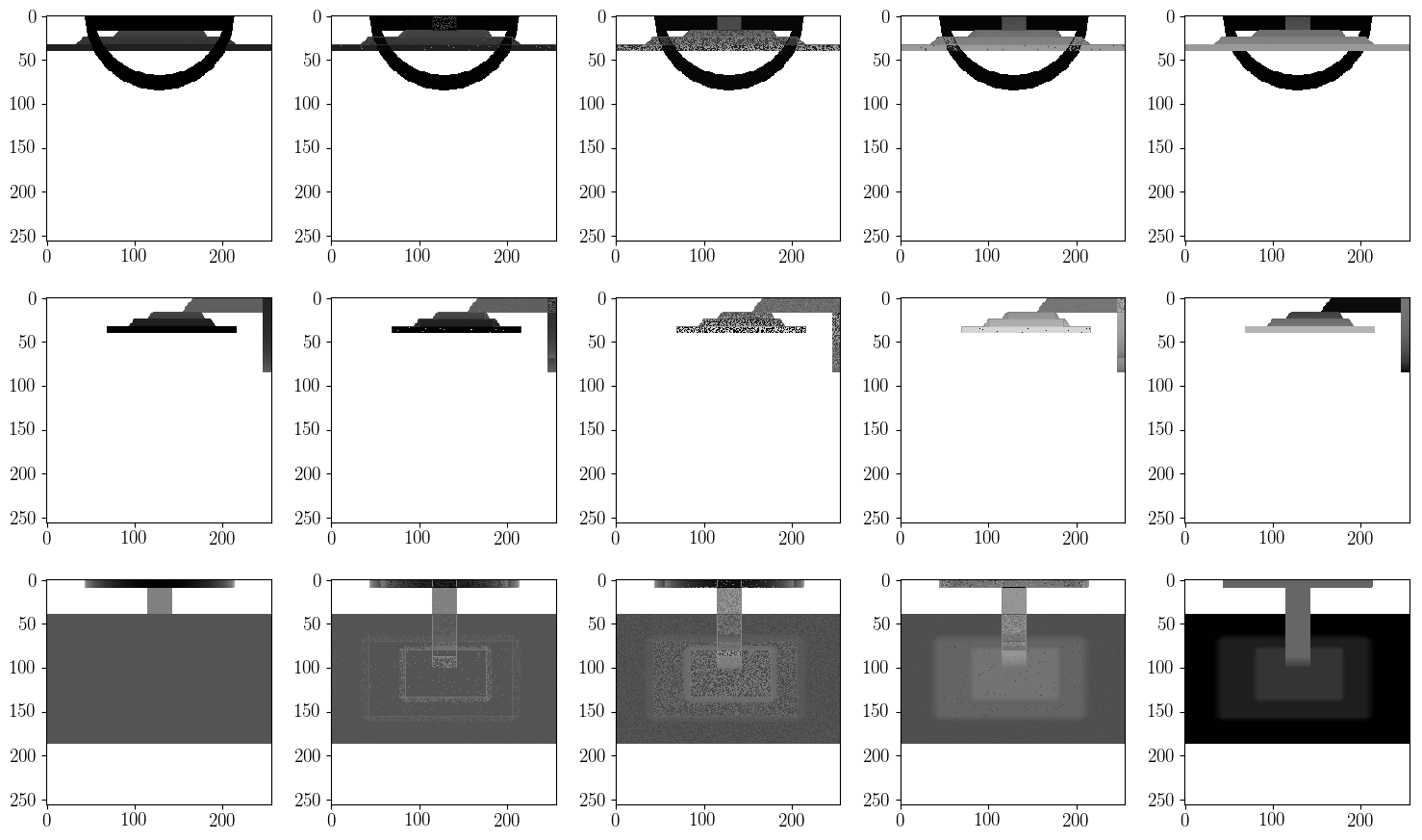}  
\end{subfigure}
\rule[1ex]{\linewidth}{0.5pt}
\begin{subfigure}{\tablfeaturea\linewidth}
  \centering
  \includegraphics[width=\linewidth]{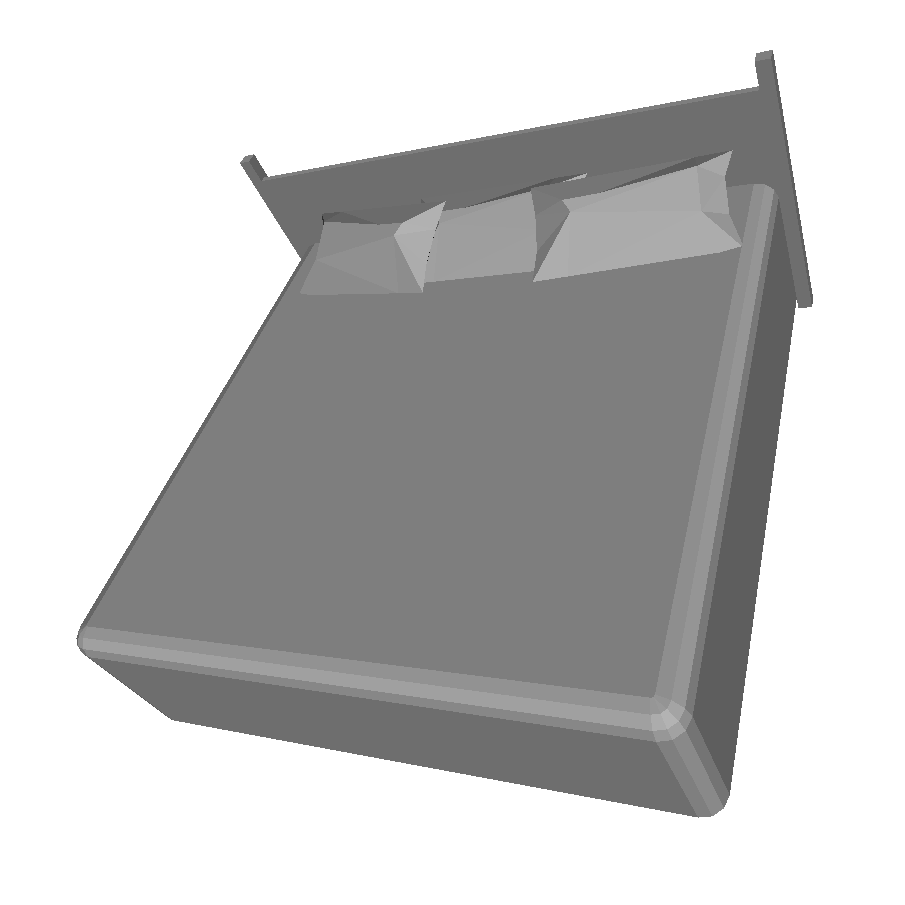}  
\end{subfigure}%
\begin{subfigure}{\tablfeatureb\linewidth}
  \centering
  \includegraphics[width=\linewidth]{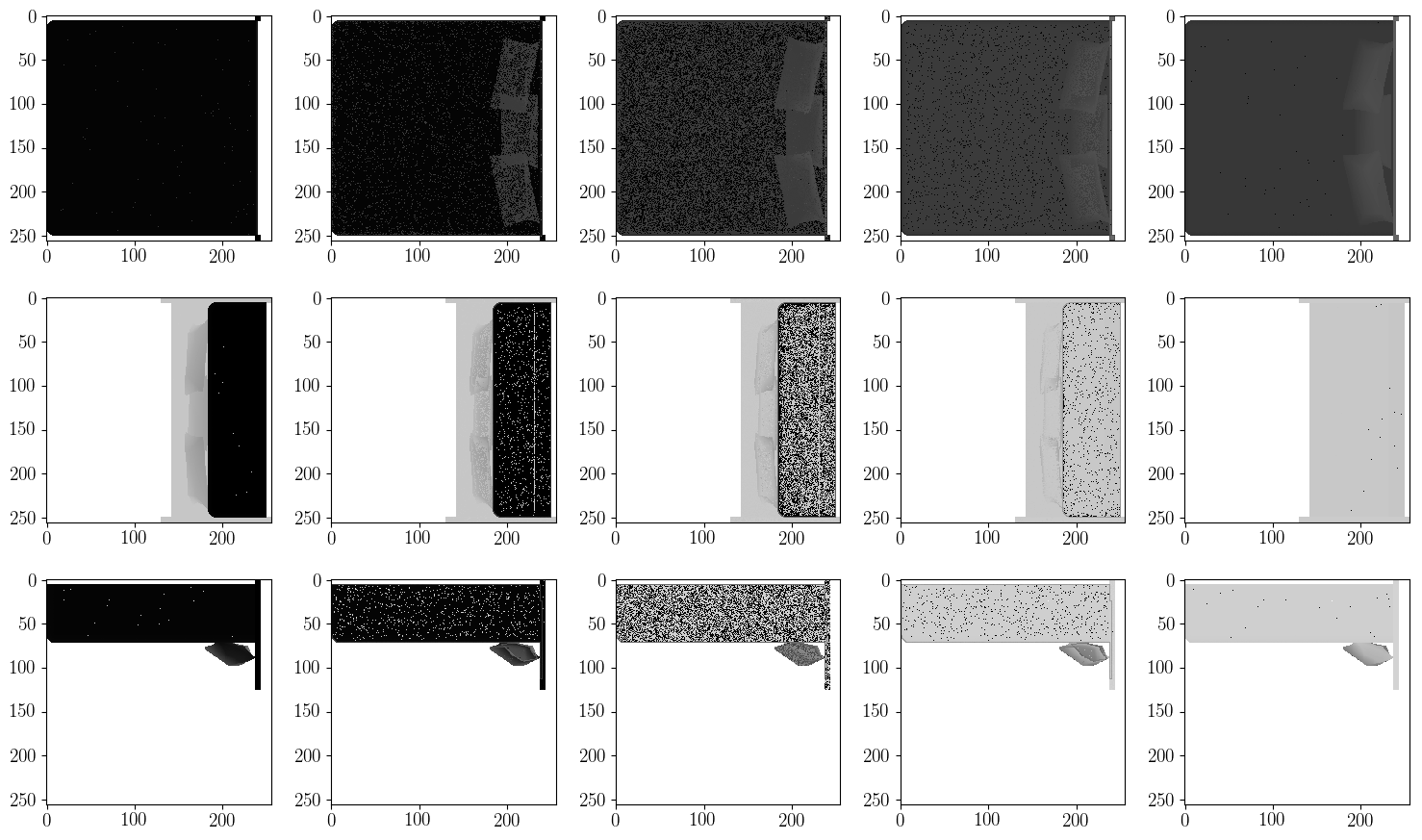}  
\end{subfigure}
\end{figure}

\begin{figure}[h!]\ContinuedFloat
\centering
\begin{subfigure}{\tablfeaturea\linewidth}
  \centering
  \includegraphics[width=\linewidth]{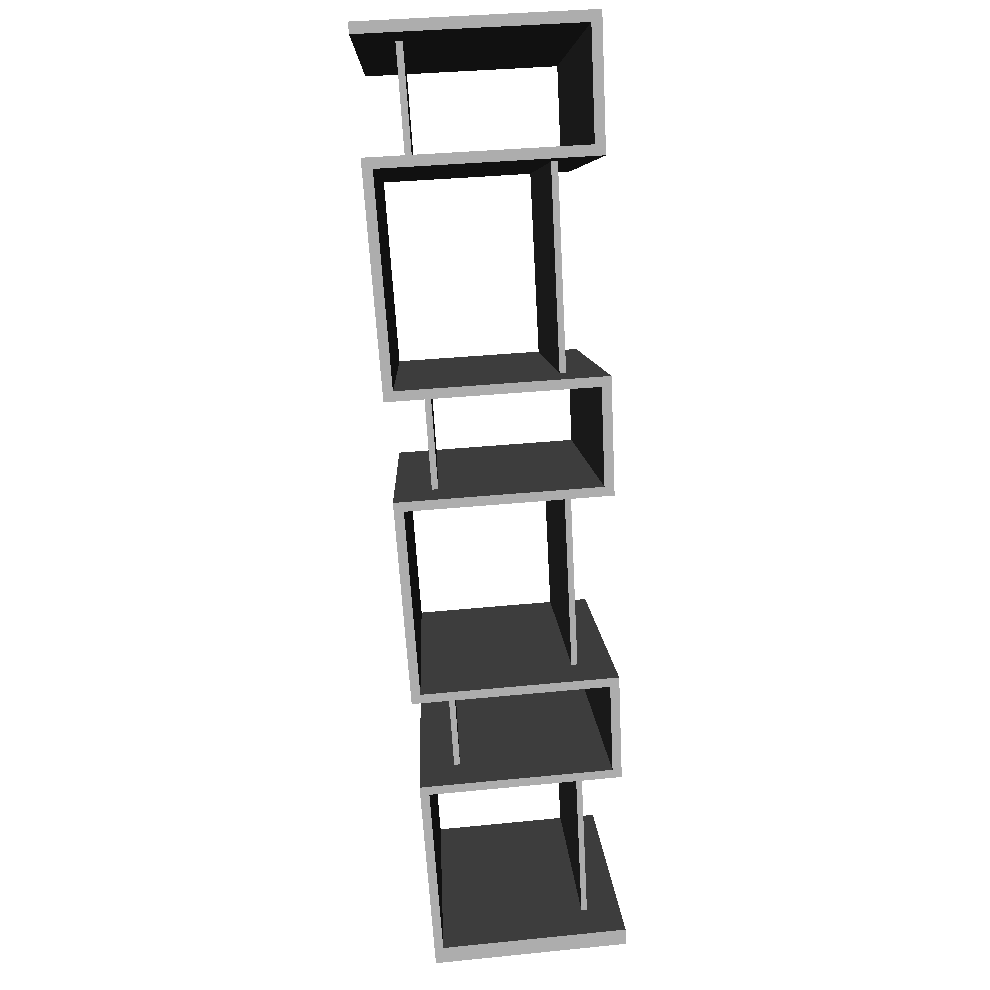}  
\end{subfigure}%
\begin{subfigure}{\tablfeatureb\linewidth}
  \centering
  \includegraphics[width=\linewidth]{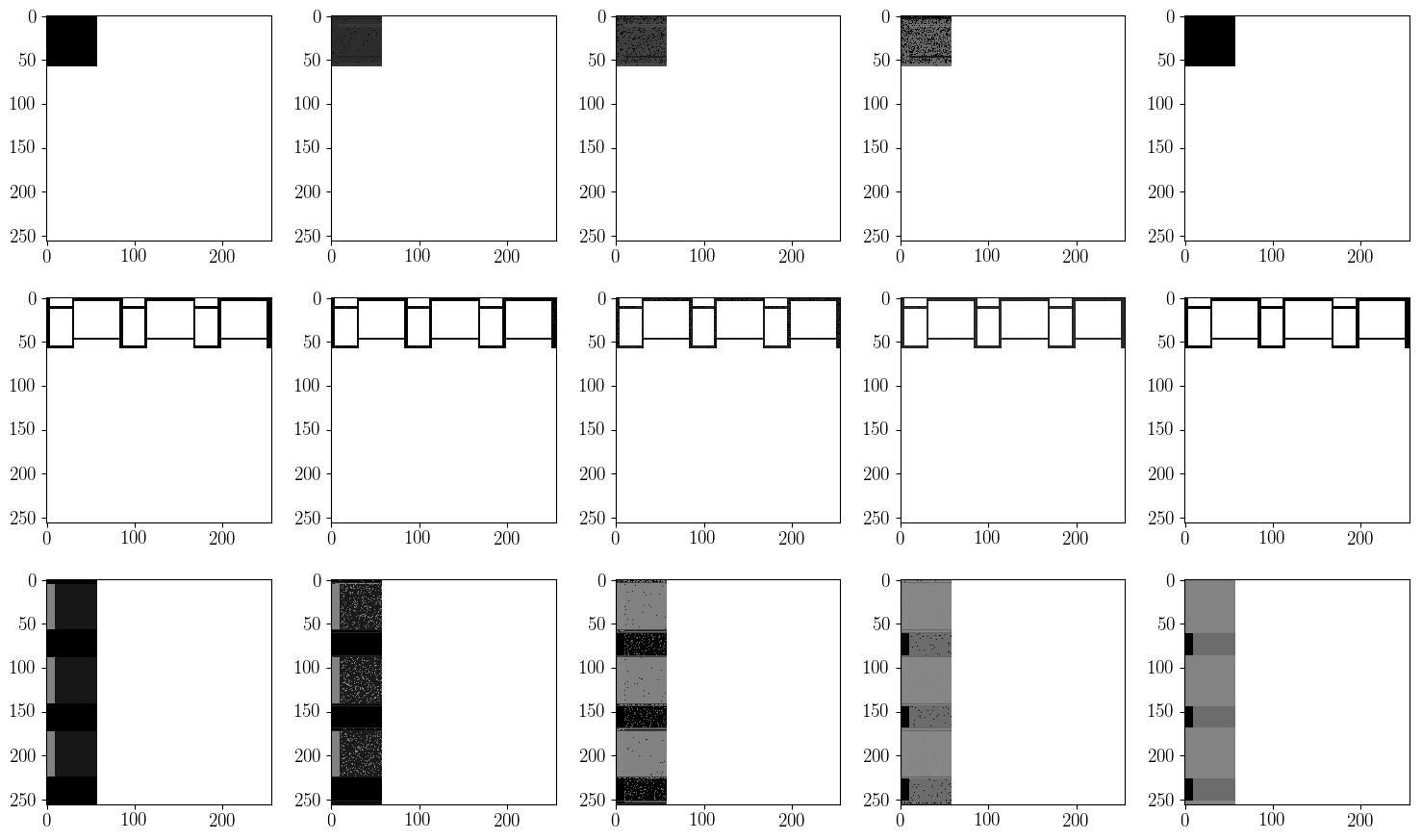}  
\end{subfigure}
\rule[1ex]{\linewidth}{0.5pt}
\begin{subfigure}{\tablfeaturea\linewidth}
  \centering
  \includegraphics[width=\linewidth]{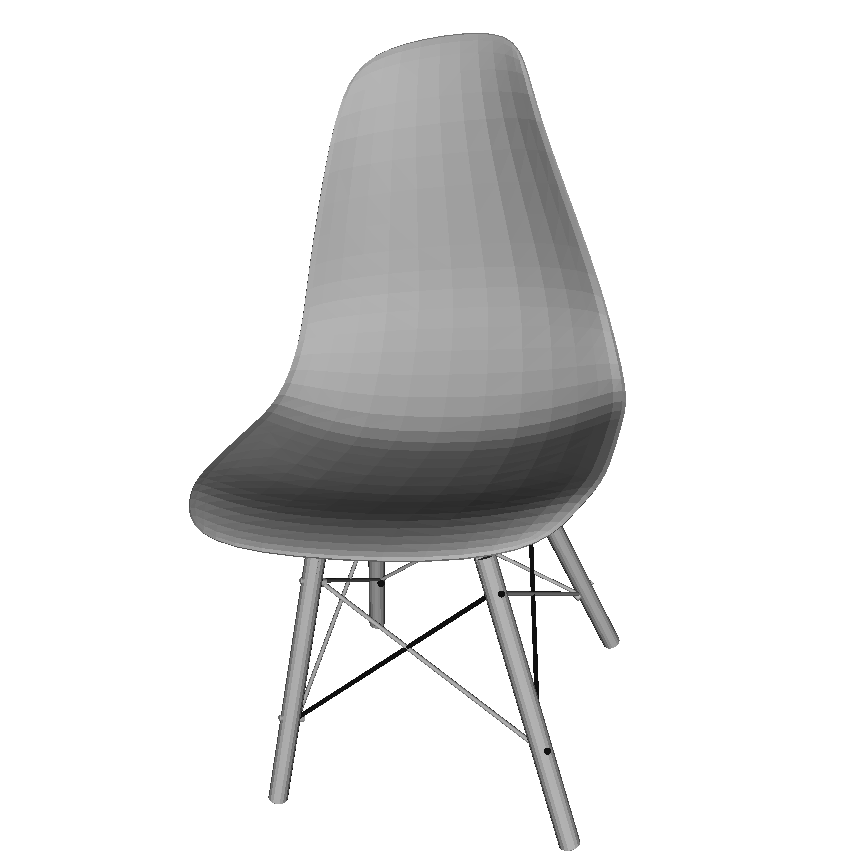}  
\end{subfigure}%
\begin{subfigure}{\tablfeatureb\linewidth}
  \centering
  \includegraphics[width=\linewidth]{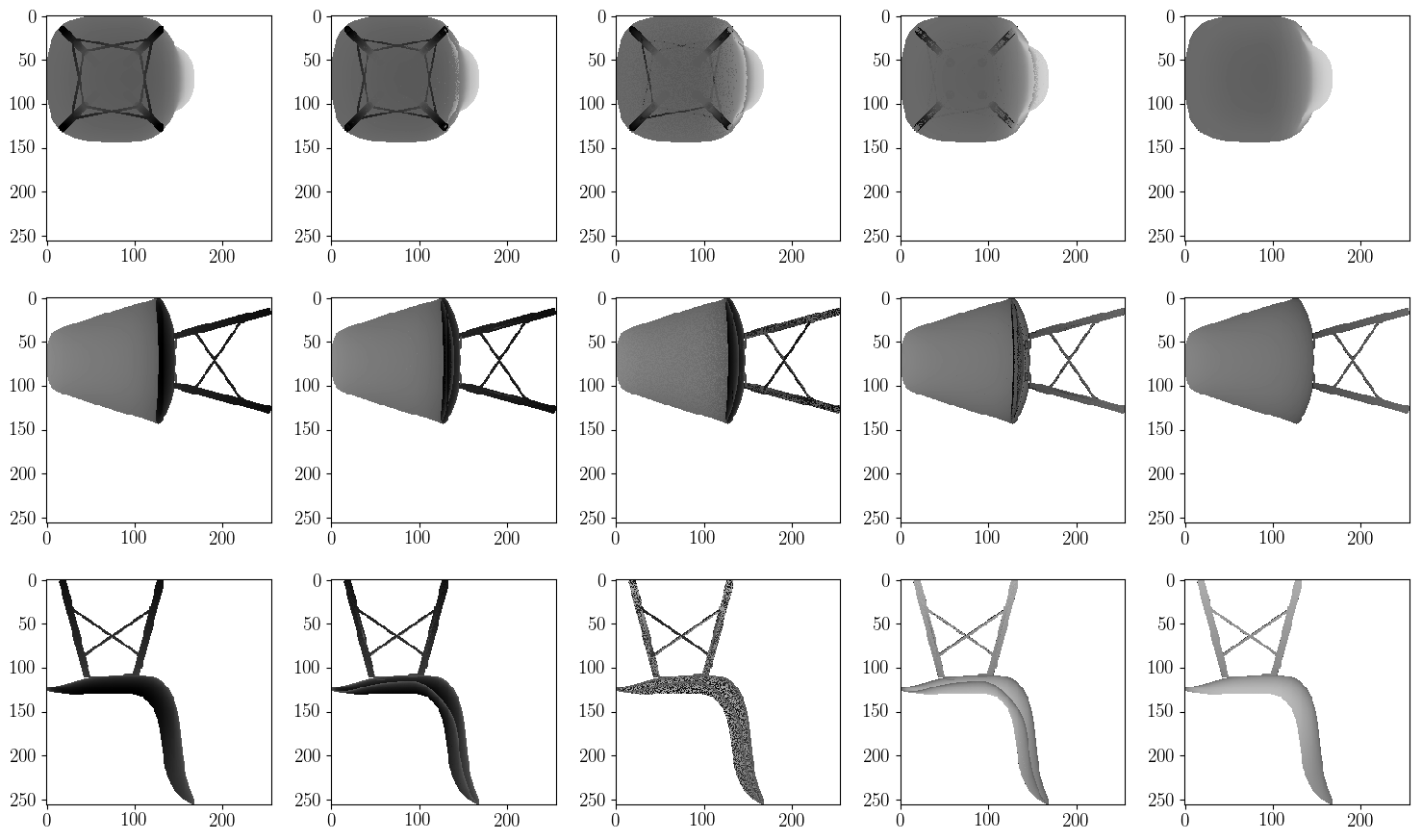}  
\end{subfigure}
\rule[1ex]{\linewidth}{0.5pt}
\begin{subfigure}{\tablfeaturea\linewidth}
  \centering
  \includegraphics[width=\linewidth]{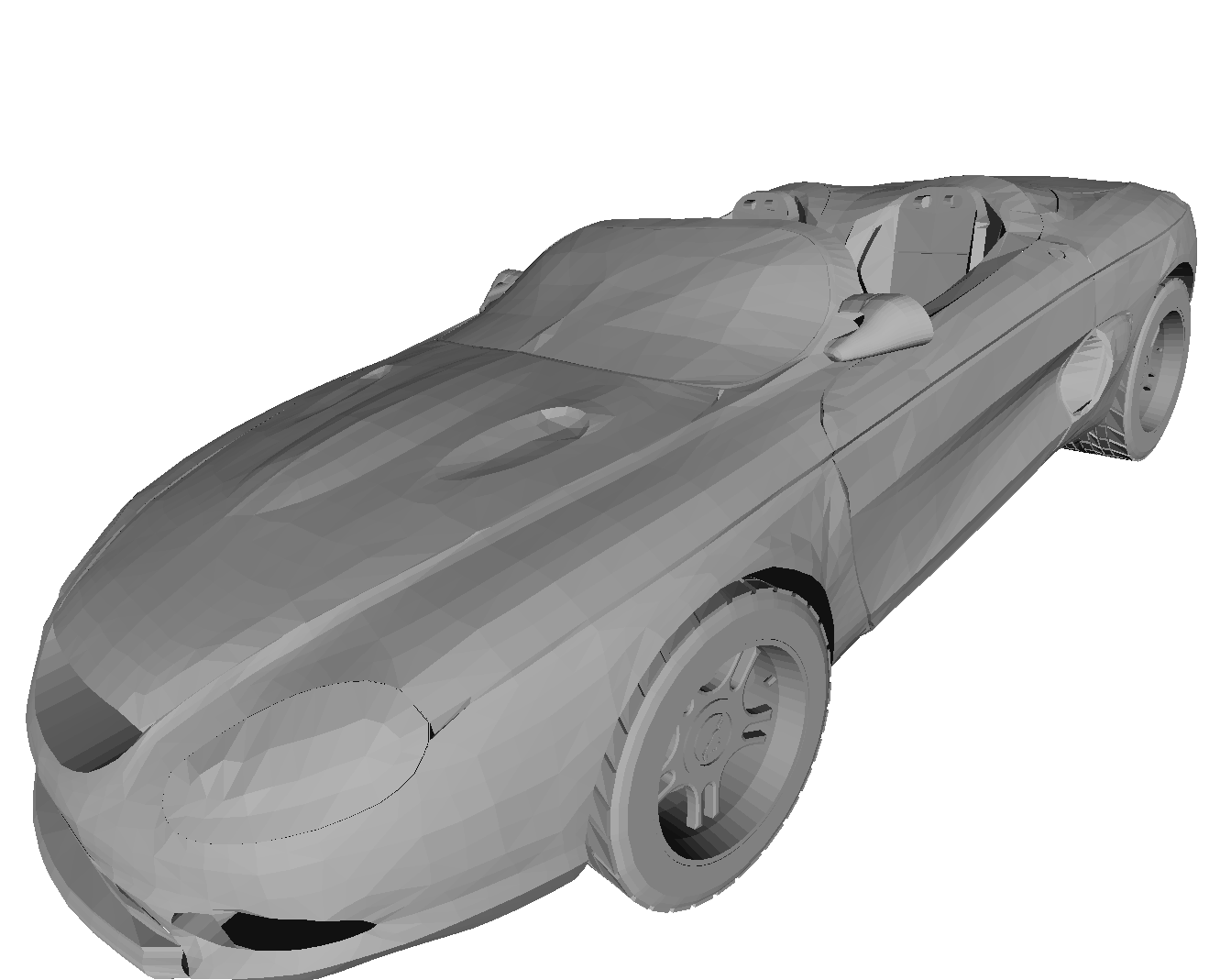}  
\end{subfigure}%
\begin{subfigure}{\tablfeatureb\linewidth}
  \centering
  \includegraphics[width=\linewidth]{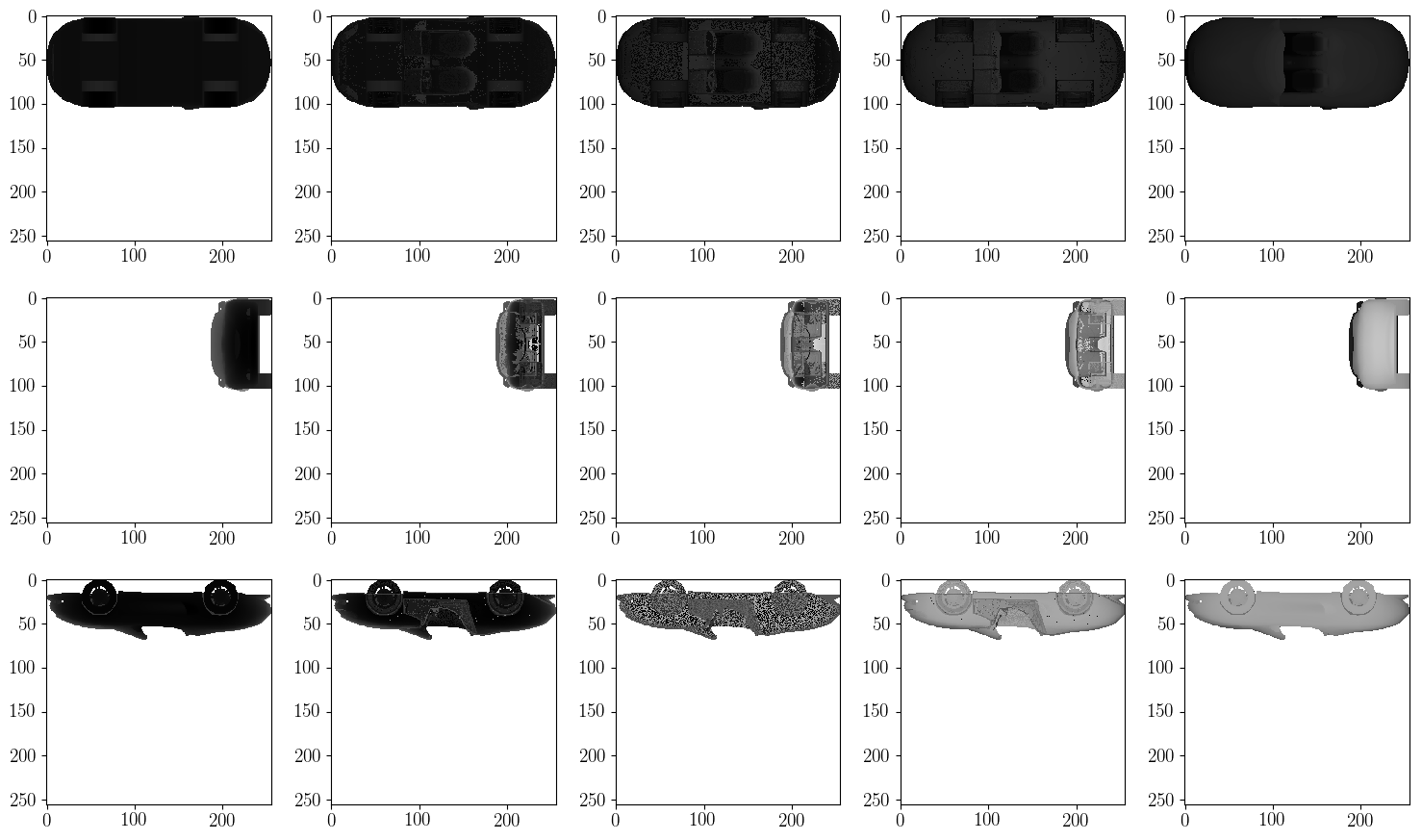}  
\end{subfigure}
\rule[1ex]{\linewidth}{0.5pt}
\caption{Visualization of the multi-layered height-map descriptors of a few shapes. Each row represents a view direction (Z, X and Y in the order). Each column represent a layer (starting from 1 to 5). Note the distinctive features captured by the different layers - specially by the 1st and 5th layer. Eg, in the Z view of the car, tyres are captured by the 1st layer, hood and the roof by the 5th layer, while the seats and interiors like seats are captured by the intermediate layers.}
\label{fig:fieature}
\end{figure}

%\begin{table}[htbp]
%\begin{center}
%\begin{tabular}{|l|}
%\hline
%convT(4x4, 512)  \\ 
%ReLU \\ \hline
%convT(4x4, 256, (2,2))  \\ 
%ReLU \\ \hline
%convT(4x4, 128, (2,2))  \\ 
%ReLU \\ \hline
%convT(4x4, 64, (2,2))  \\ 
%ReLU \\ \hline
%convT(4x4, 5, (2,2))  \\
%Tanh \\ \hline
%\end{tabular}
%\end{center}
%\caption{}
%\label{}
%\end{table}
%
%\begin{table}[htbp]
%\begin{center}
%\begin{tabular}{|l|}
%\hline
%conv(4x4, 64, (2,2)) \\ \hline
%LReLU(0.2) \\ \hline
%conv(4x4, 128, (2,2)) \\ \hline
%LReLU(0.2) \\ \hline
%conv(4x4, 256, (2,2)) \\ \hline
%LReLU(0.2) \\ \hline
%conv(4x4, 512, (2,2)) \\ \hline
%\end{tabular}
%\end{center}
%\caption{}
%\label{}
%\end{table}
\begin{table}[]
\begin{center}
\begin{tabular}{|l|c|r|r|}
\hline
Network Architecture & Activation olume & Volume size&Memory \\ \hline
\textit{input} & 256x256x256x1 & 16777216&67108864 \\ \hline
conv(1, 8) & 256x256x256x8 & 134217728&536870912 \\ 
conv(8, 14) & 256x256x256x14 & 234881024&939524096 \\ 
maxpool(2) & 128x128x128x14 & 29360128&117440512 \\ \hline
conv(14, 14) & 128x128x128x14 & 29360128&117440512 \\ 
conv(14, 20) & 128x128x128x20 & 41943040&167772160 \\ 
maxpool(2) & 64x64x64x20 & 5242880&20971520 \\ \hline
conv(20, 20) & 64x64x64x20 & 5242880&20971520 \\ 
conv(20, 26) & 64x64x64x26 & 6815744&27262976 \\ 
maxpool(2) & 32x32x32x26 & 851968&3407872 \\ \hline
conv(26, 26) & 32x32x32x26 & 851968&3407872 \\ 
conv(26, 32) & 32x32x32x32 & 1048576&4194304 \\ 
maxpool(2) & 16x16x16x32 & 131072&524288 \\ \hline
conv(32, 32) & 16x16x16x32 & 131072&524288 \\ 
conv(32, 32) & 16x16x16x32 & 131072&524288 \\ 
maxpool(2) & 8x8x8x32 & 16384&65536 \\ \hline
\textit{\textbf{Total} } &  & 507002880&2028011520 \\ 
 &  &&{$\approx$ 1.89 GB} \\ \hline
\end{tabular}
\end{center}
\caption{Memory computation for DenseNet256\cite{Riegler2017a} for one sample. For a batch size of 32 we get 32*1.8 $\approx$ 60GB of memory (\textit{Table 3, Main Paper}). This value is also verified against the values in the plot provided in Figure 7(a) \cite{Riegler2017a} }
\label{table:octnet256}
\end{table}

\begin{table}[tbp]

\begin{center}
\begin{tabular}{|l|c|r|r|}
\hline
Network Architecture & Activation Volume & \multicolumn{1}{l|}{Volume size} &  Memory\\ \hline
input & 256x256x5 & 327680 &   1310725 \\ \hline
conv(5,64) & 256x256x64 & 4194304 &   16777221 \\ 
conv(64,64) & 256x256x64 & 4194304 &   16777221 \\ 
maxpool(2) & 128x128x64 & 1048576 &   4194309 \\ \hline
conv(64,128) & 128x128x128 & 2097152 &   8388613 \\ 
conv(128,128) & 128x128x128 & 2097152 &   8388613 \\ 
maxpool(2) & 64x64x128 & 524288 &   2097157 \\ \hline
conv(128,256) & 64x64x256 & 1048576 &   4194309 \\ 
conv(256,256) & 64x64x256 & 1048576 &   4194309 \\ 
conv(256,256) & 64x64x256 & 1048576 &   4194309 \\ 
maxpool(2) & 32x32x256 & 262144 &   1048581 \\ \hline
conv(256,512) & 32x32x512 & 524288 &   2097157 \\ 
conv(512,512) & 32x32x512 & 524288 &   2097157 \\ 
conv(512,512) & 32x32x512 & 524288 &   2097157 \\ 
maxpool(2) & 16x16x512 & 131072 &   524293 \\ \hline
conv(512,512) & 16x16x512 & 131072 &   524293 \\ 
conv(512,512) & 16x16x512 & 131072 &   524293 \\ 
conv(512,512) & 16x16x512 & 131072 &   524293 \\ 
maxpool(2) & 8x8x512 & 32768 &  131077 \\ \hline
\textit{Total (branch)} &  & \multicolumn{1}{l|}{} &   80084992 $\approx$ 80 MB \\ \hline
FC1 & 1x1x4096 & 4096 &   16389 \\ \hline
FC2 & 1x1x4096 & 4096 &   16389 \\ \hline
FC3 & 1x1x40 & 40 &   165 \\ \hline
\textit{\textbf{Total}} &&& $\approx$ 112 MB. \\ \hline
\end{tabular}
\end{center}
\caption{Memory computation for Our single view net using VGG  for one sample. For a batch size of 32 we get 32*112 $\approx$ 3.5GB of memory. But our experiments which was run using a PyTorch implementation for a batchsize of 32 occupies 8 GB of data in GeForce GTX 1080 Ti. We used this relaxed value of 8GB in \textit{Table 3 (Right), Main Paper} for a more fare comparison against OctNet\cite{Riegler2017a}. For OctNet we used the memory consumed in a similar settings using their representation and varified its value in the plot provided in the Figure 7(a) of \cite{Riegler2017a}.}
\label{table:singleview}
\end{table}

\begin{table}[htbp]
\begin{center}
\begin{tabular}[b]{|l|c|}
\hline
\textit{input} - 100x1 \\ \hline
convT(4x4, 512)  \\ 
ReLU \\ \hline
convT(4x4, 256, (2,2))  \\ 
ReLU \\ \hline
convT(4x4, 128, (2,2))  \\ 
ReLU \\ \hline
convT(4x4, 64, (2,2))  \\ 
ReLU \\ \hline
convT(4x4, 5, (2,2))  \\
Tanh \\ \hline
\textit{output} - 64x64x5 \\ \hline 
\end{tabular}%
\qquad \qquad \qquad
\begin{tabular}[b]{|l|}
\hline
\textit{input} - 64x64x5 \\ \hline
conv(4x4, 64, (2,2)) \\ 
LReLU(0.2) \\ \hline
conv(4x4, 128, (2,2)) \\ 
LReLU(0.2) \\ \hline
conv(4x4, 256, (2,2)) \\ 
LReLU(0.2) \\ \hline
conv(4x4, 512, (2,2)) \\ \hline
\textit{output} - 8x8x512 \\ \hline

\end{tabular}
\end{center}
\caption{(Left) The generative branch of the Multiview DCGAN which produces MLH vector of size 64x64x5. (Right) The discriminator part of the MV DCGAN which takes the generated 64x64x5 input and produces an activation volume of 8*8*512. In the multiview design we have 3 independent generator branches and 3 independent discriminator branches. The 3 output volume of the discriminator of the discriminator is concatenated by a non-commutative operation followed by FC(1024) and FC(1). More details are in \textit{Figure 4, Main paper}. The numbers in the bracket (x,x) denote the stride values for the strided convolution and transposed convolution blocks. Batch-normalization is used between every layers except for the first and the last layers.}
\label{table:gan}
\end{table}

\begin{table}
\centering
\begin{tabular}{|m{0.2\linewidth}|m{0.2\linewidth}|m{0.25\linewidth}|m{0.25\linewidth}|}
\hline
\centering GT label & Predicted label & \centering Misclassified &  Sample from \\ 
& & \centering shape &  predicted label \\ \hline
&&& \\
\centering plant & \centering flower pot & 
\includegraphics[width=\linewidth]{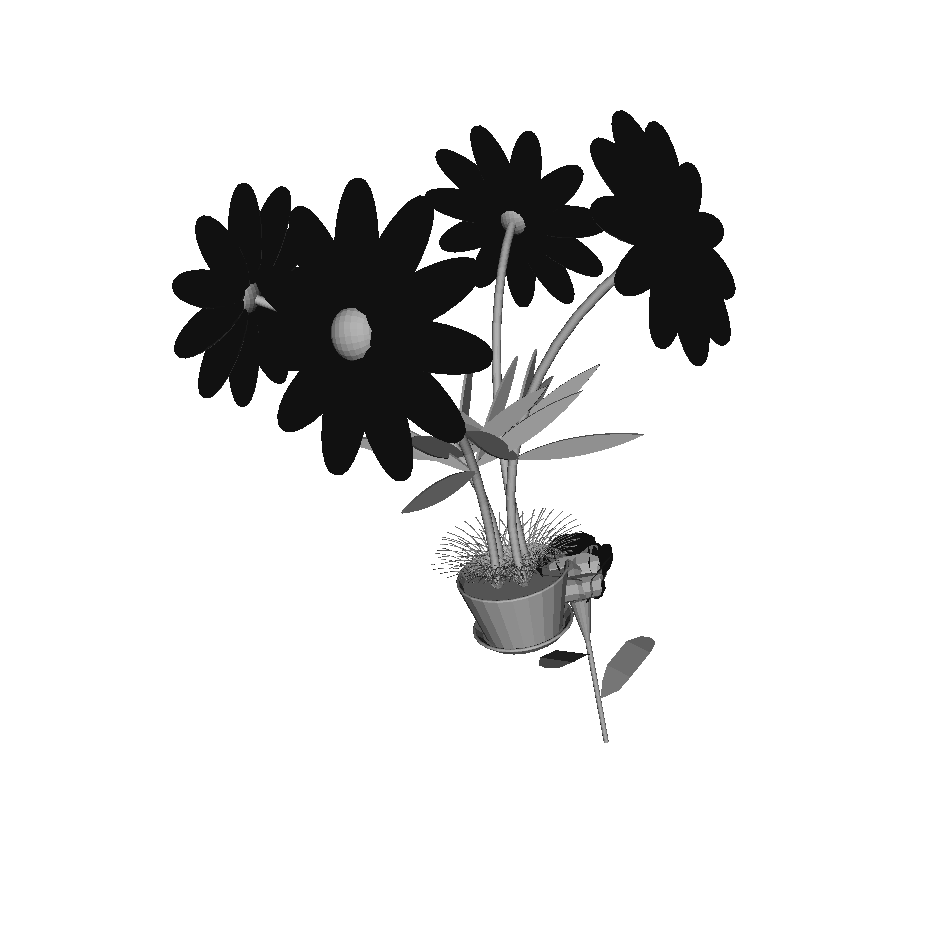}&
\includegraphics[width=\linewidth]{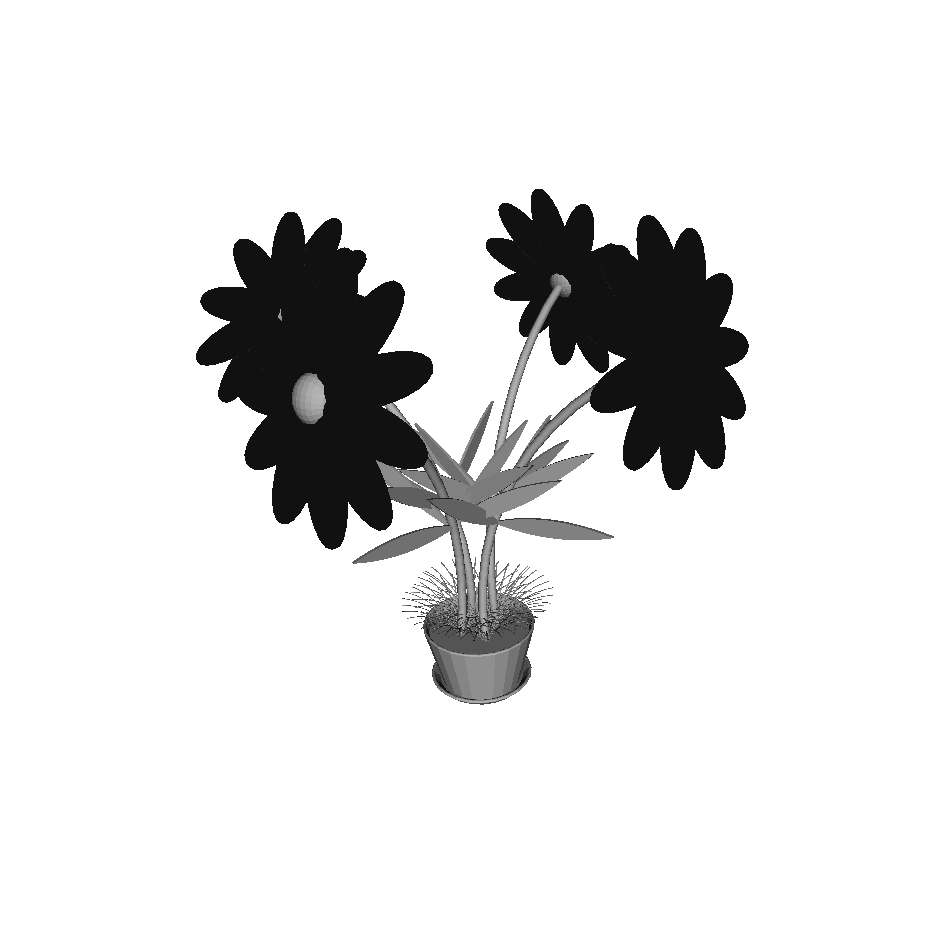}\\

\centering vase & \centering cup pot & 
\includegraphics[width=\linewidth]{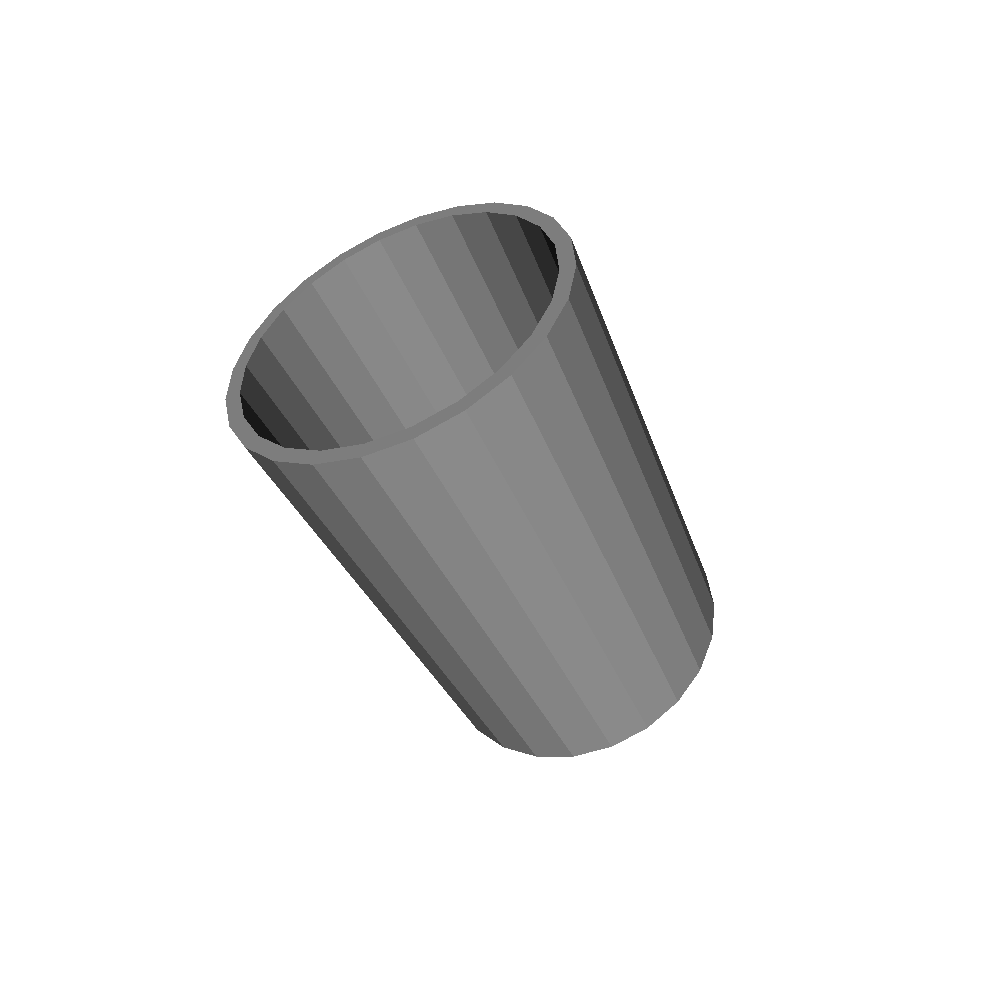}&
\includegraphics[width=0.8\linewidth]{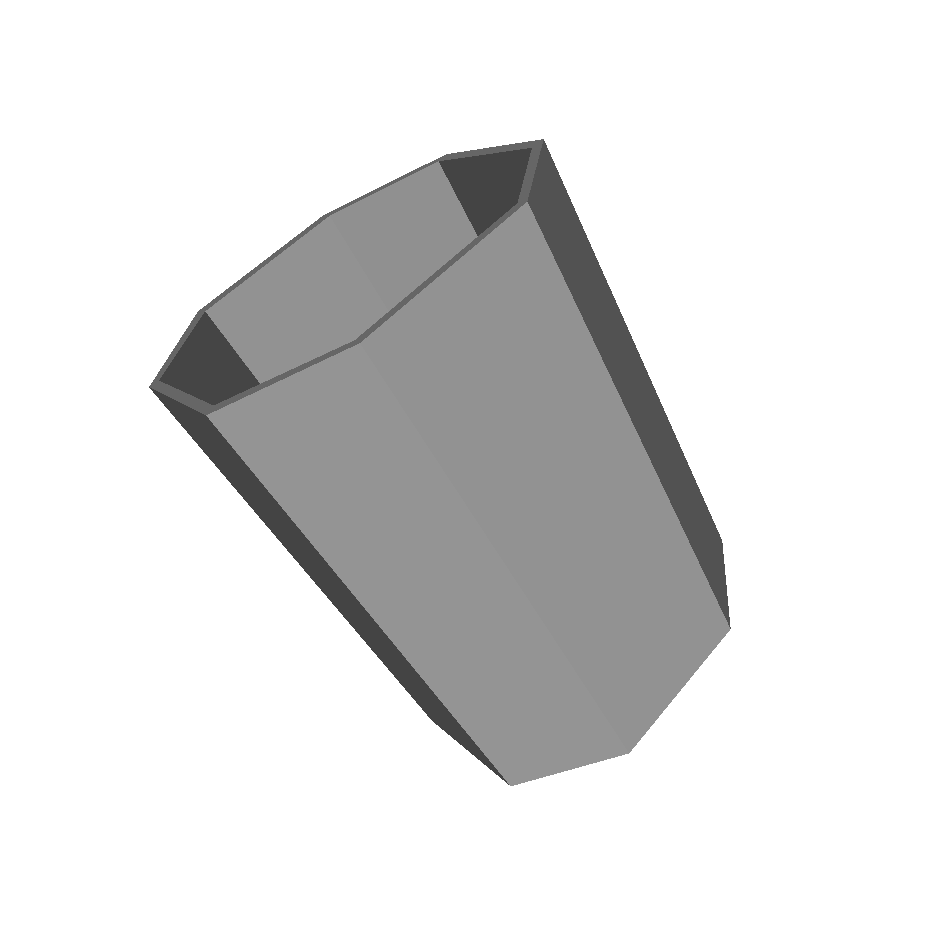}\\

\centering desk & \centering table & 
\includegraphics[width=\linewidth]{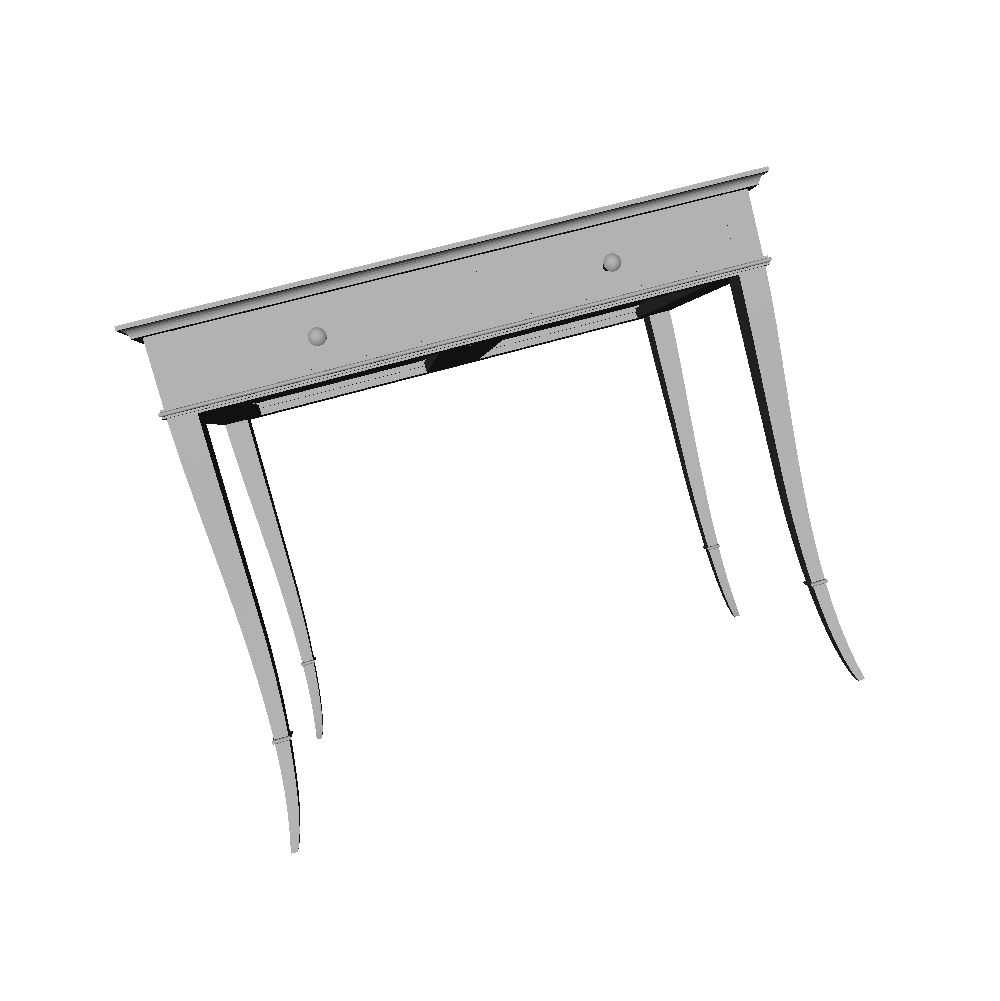}&
\includegraphics[width=0.9\linewidth]{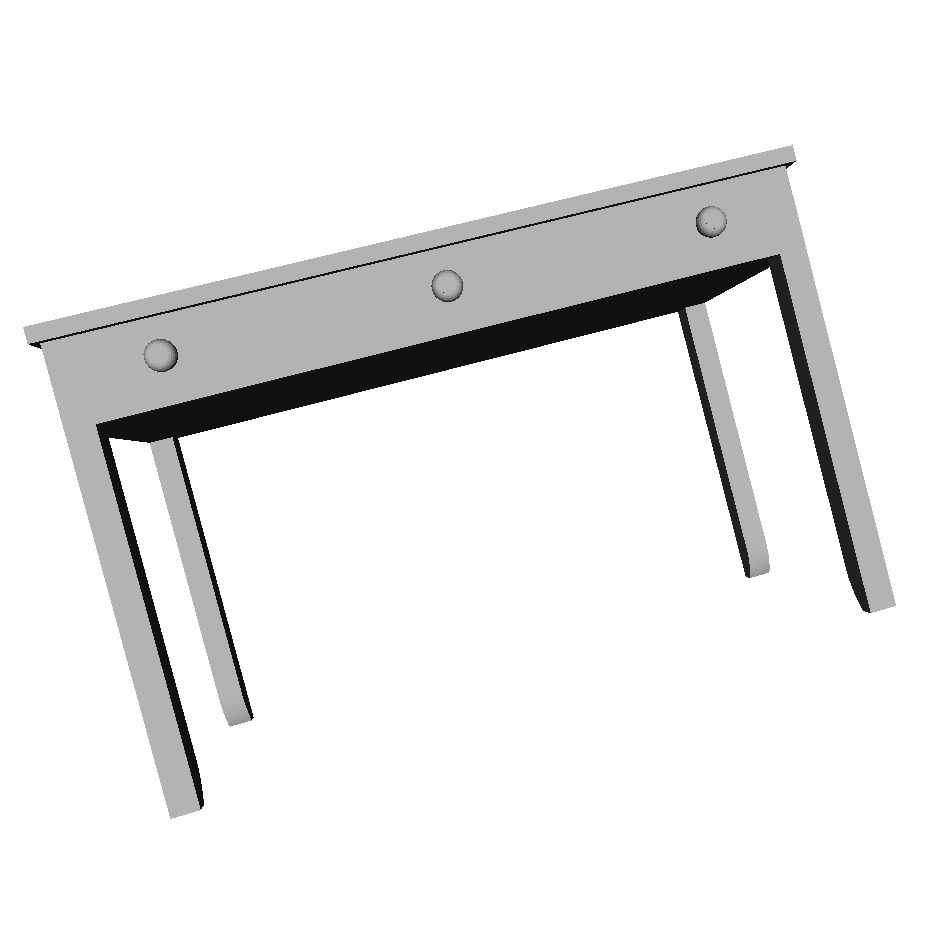}\\

\centering night stand & \centering dresser & 
\includegraphics[width=\linewidth]{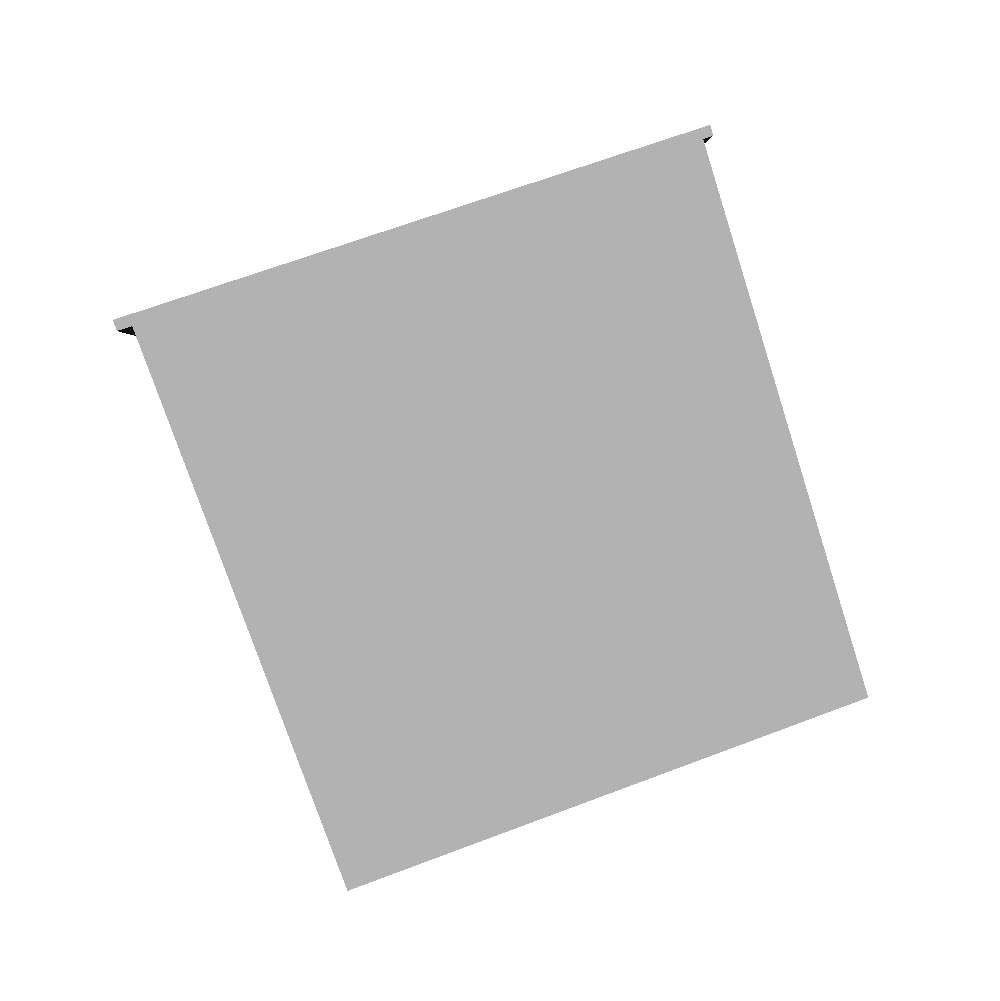}&
\includegraphics[width=0.8\linewidth]{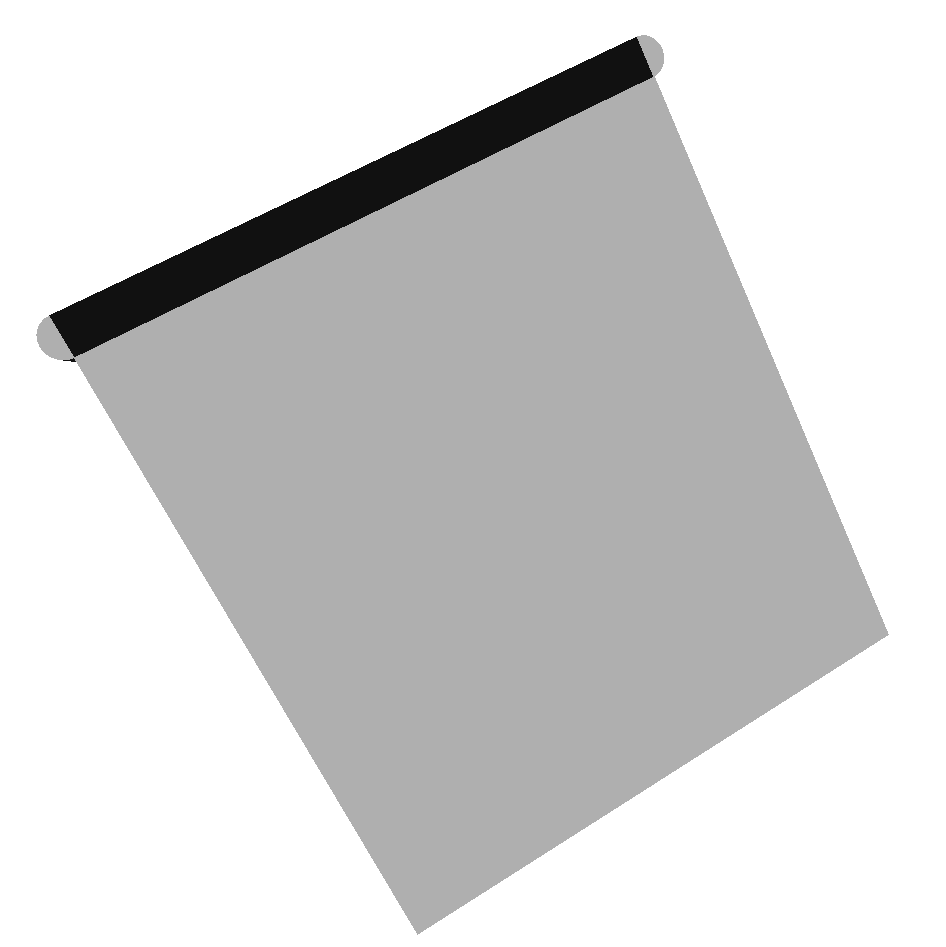}\\

\centering table & \centering desk & 
\includegraphics[width=0.9\linewidth]{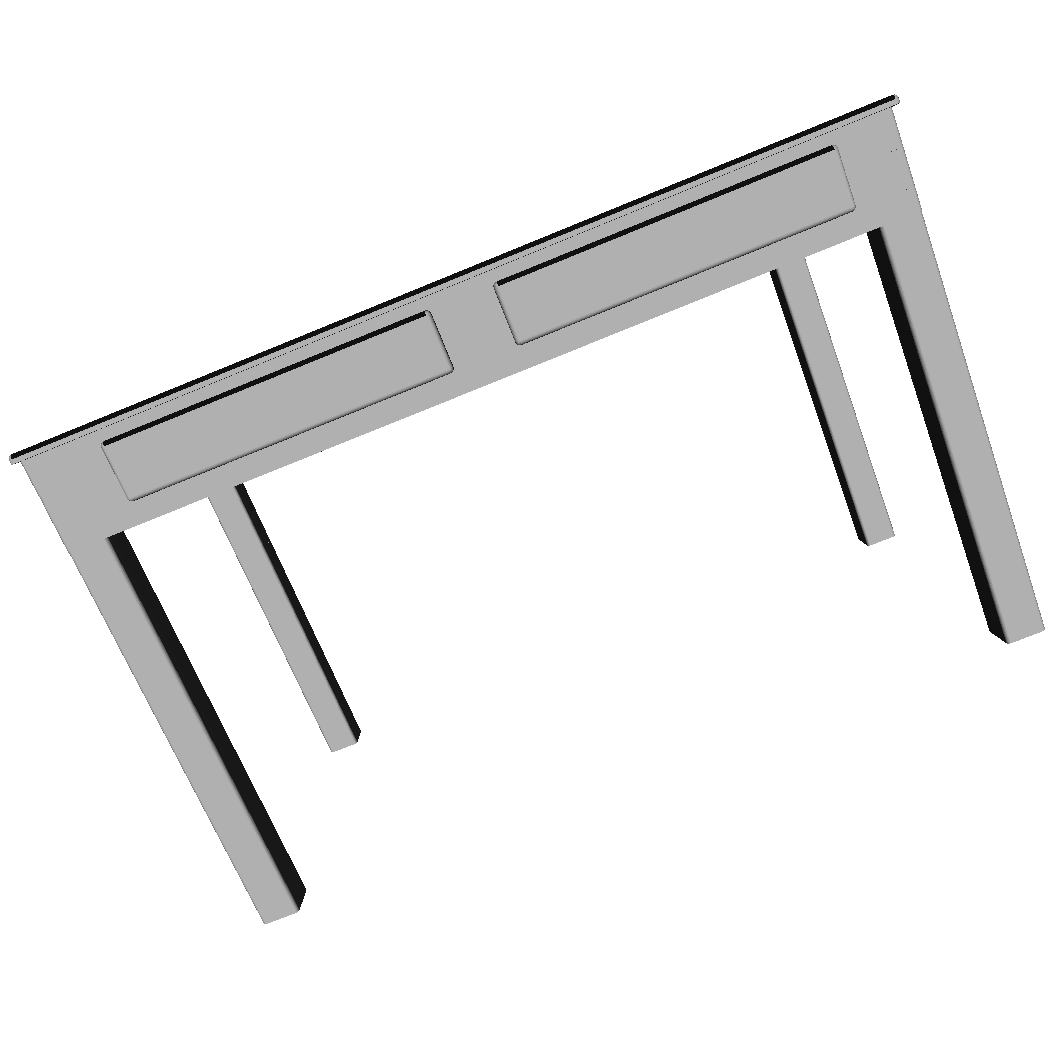}&
\includegraphics[width=\linewidth]{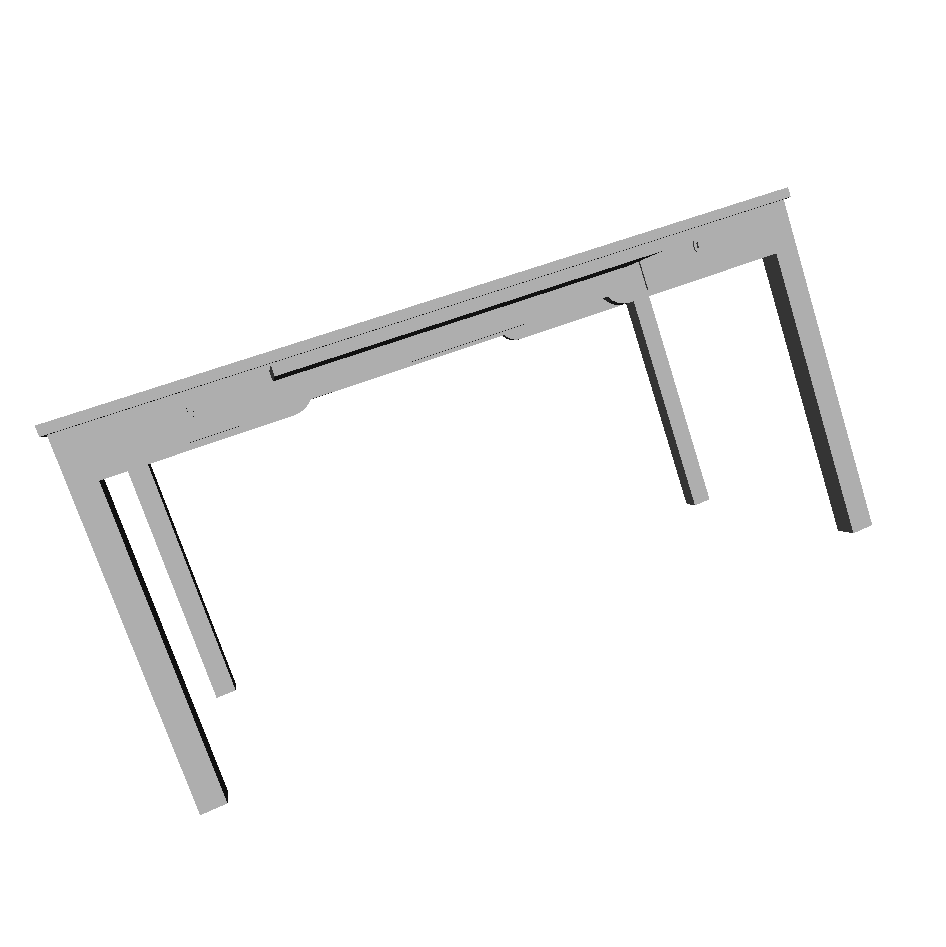}\\

\hline
\end{tabular}
\caption{Example of misclassified shape and a close looking sample from its predicted label in ModelNet40.}
\label{table:miscl}
\end{table}

\end{document}